\pdfoutput=1

\documentclass[11pt,UTF8]{article}

\usepackage[preprint]{acl}

\usepackage{times}
\usepackage{latexsym}

\usepackage[T1]{fontenc}

\usepackage[utf8]{inputenc}

\usepackage{microtype}

\usepackage{inconsolata}

\usepackage{graphicx}
\usepackage{algcompatible}
\usepackage{algorithm}
\usepackage{amsthm}
\usepackage{amsmath}
\usepackage{booktabs}
\usepackage{amssymb}
\usepackage{graphicx}
\usepackage{subcaption}
\usepackage{multirow}
\usepackage{tabularx}
\usepackage{diagbox}
\usepackage{color}
\usepackage{xcolor}
\usepackage{hyperref}
\usepackage{CJK}
\usepackage{listings}
\usepackage{fancyvrb}
\title{Can MLLMs Generalize to Multi-Party dialog? Exploring Multilingual Response Generation in Complex Scenarios}

\author{
  \textbf{Zhongtian Hu\textsuperscript{1}},
  \textbf{Yiwen Cui\textsuperscript{1}},
  \textbf{Ronghan Li\textsuperscript{2}},
  \textbf{Meng Zhao\textsuperscript{3}},
  \textbf{Lifang Wang\textsuperscript{1}},
\\
  \textsuperscript{1}School of Computer Science and Engineering, Northwestern Polytechnical University,
  \\\textsuperscript{2}School of Computer Science and Technology, Xidian University,
  \\\textsuperscript{3}School of Artificial Intelligence and Big Data, Henan University of Technology
\\
	\texttt{ahxchzt@mail.nwpu.edu.cn},
	\texttt{wanglf@nwpu.edu.cn} 
  }

\begin{document}
	\maketitle
\begin{abstract}
	Current multilingual large language models(MLLMs) still focus on simple question-answering formats, often overlooking more complex dialogue scenarios. In other words, their capabilities of multilingual large models have yet to be validated in dialogue tasks with intricate structures. We therefore ask, \emph{Q1:} How well do LLMs generalize to more complex dialog scenarios? \emph{Q2:} Can supervised fine-tuning on a high-quality parallel benchmark restore this ability? \emph{Q3:} Does the "multilingual complementarity" effect survive in the setting? To answer these questions, we introduce \textbf{XMP}\footnote{https://anonymous.4open.science/r/XMP\_data-5CC7}, a high-quality parallel Multilingual(\textbf{X}) dataset sourced from \textbf{M}ulti-party \textbf{P}odcast dialogues, which is the first parallel dataset focusing on multi-party dialogue scenarios. Most samples in the dataset feature three or more participants, discussing a wide range of topics. Through extensive experiments, we find that, \emph{R1:} MLLMs fail to generalize to multi-party setting, \emph{R2:} Fine-tuning on XMP improves only marginally, with the 70B model achieving at most a 1\% absolute gain over its 8B counterpart; \emph{R3:} Mixing languages during SFT is usually detrimental, with any benefits being marginal and limited to isolated cases in the 70B model.
\end{abstract}
\section{Introduction}
Amid the rising demand for seamless cross-lingual communication and the growing complexity of global language needs, multilingual research has garnered widespread attention \cite{xue-etal-2021-mt5,liu-etal-2020-multilingual-denoising,conneau-etal-2020-unsupervised}. Expectations for large language models (LLMs) have grown substantially, \citet{touvron2023llama,yang2024qwen2technicalreport} particularly their ability to handle a wide range of multilingual NLP tasks \cite{lin2021xpersona,liu2023xdailydialog,ponti2020xcopa,muennighoff2022crosslingual}. Among these tasks, dialogue generation stands out as one of the most crucial. However, current LLMs have only demonstrated their multilingual capabilities in relatively simple, two-party dialogue scenarios. More complex interaction settings, such as multinational meetings or multi-player role-playing games are largely unexplored.
\begin{figure}[!t]
	\centering
	\resizebox{\linewidth}{!}{ 
		\includegraphics[width=\textwidth]{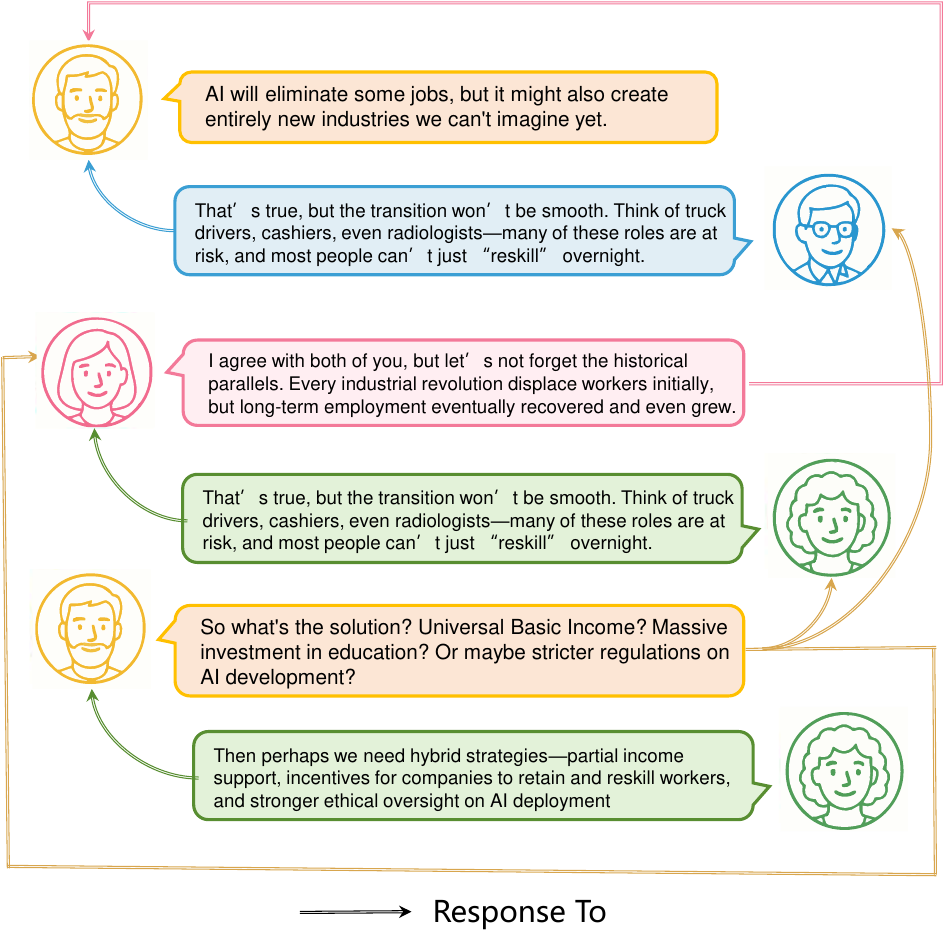}}
	\caption{An example of a multi-party dialogue illustrating complex conversational dependencies, showcasing turn-taking, contextual references, and opinion shifts.}
	\label{fig:example_dialogue}
\end{figure}

To date, no multilingual dataset exists that focuses specifically on more complex dialogue settings involving long contexts and multiple speakers \cite{hu2019gsn}. An example of such a scenario is illustrated in Figure \ref{fig:example_dialogue}, where the structure and dynamics of the conversation highlight its complexity. This \emph{absence of suitable datasets creates a critical research gap}, severely limiting our understanding of how effectively existing MLLMs generalize to realistic and complex dialogue scenario.

We systematically address the following clearly defined research questions, \emph{Q1:} How well do LLMs generalize to more complex dialog scenarios? \emph{Q2:} Can supervised fine-tuning on a high-quality parallel benchmark restore this ability? \emph{Q3:} Prior studies like \cite{mulcaire2019polyglot,conneau2020emerging,whitehouse2023llm} have shown that mixing multiple languages during training can enhance model performance. Does this effect still hold under the more complex dialogue settings considered in our work?

To investigate these issues, we present XMP, a high-quality parallel multilingual dataset for multi-party dialogue. The dataset is built from real-world multi-speaker interactions in podcasts, and the original English data has been carefully translated into five languages:  Chinese, Japanese, German, French, and Italian. The dataset consists of 312k conversations spanning diverse topics such as society, culture, politics, and entertainment. To ensure translation quality, we implemented a rigorous quality control process, including systematic data filtering and cleaning to maintain the diversity and semantic coherence of the dialogue content. In addition, we conducted a comprehensive quality assessment to ensure both cross-lingual consistency and cultural fidelity across all translated dialogues.

Through extensive experiments, we found that several previously recognized multilingual capabilities \cite{doddapaneni2021primer,wang2019cross,whitehouse2023llm} of LLMs no longer hold true on this dataset. Specifically, \emph{R1:} MLLMs exhibit significant limitations in generalizing to complex dialogue scenarios; \emph{R2:} Fine-tuning on XMP improves only marginally, with the 70B model achieving at most a 1\% absolute gain over its 8B counterpart; \emph{R3:} The multilingual complementarity effect largely diminishes in complex, multi-party dialogues. Our key contributions are summarized as follows:
\begin{itemize}
	\item We construct and release XMP, the first high-quality parallel multilingual dataset for multi-party dialogue, covering six languages and over 312k conversations. The dataset offers diverse topics and supports research into multilingual multi-party conversation and cross-lingual transfer.
	\item We conduct extensive evaluations on XMP, revealing significant limitations in the multilingual capabilities of current LLMs.
	\item We reveal key insights into the performance of MLLMs, providing valuable guidance for advancing multilingual dialogue systems in realistic and challenging conversational contexts.
\end{itemize}

\section{Related Work}
\subsection{Multilingual Datasets for Dialogue}
High-quality multilingual datasets play a crucial role in advancing multilingual NLP research, such as XCOPA \cite{ponti2020xcopa} and XWinograd \cite{muennighoff2022crosslingual} primarily focus on structured tasks like multiple-choice question answering and reasoning. Although there exist multilingual dialogue datasets such as XDailyDialog \cite{liu2023xdailydialog} and XPersona \cite{lin2021xpersona}, these datasets primarily emphasize straightforward two-party dialogues, typically lack complexity, rarely addressing multi-party scenarios, and frequently neglect the critical task of response generation. Given the fundamental importance of generating coherent and contextually relevant dialogue responses, this omission significantly restricts progress in multilingual dialogue system development.
\subsection{Dialogue in Complex Scenarios}

While dialogue generation has been extensively studied, most existing benchmarks focus on \textbf{dyadic (two-party) conversations}, which significantly limits their applicability to more realistic, multi-party scenarios.

\textbf{Friends} \cite{shmueli2019socialnlp} is one of the most commonly used multi-party dialogue datasets, but it suffers from a key limitation: it includes only six fixed speakers from scripted television show dialogues. This restricts its generalizability, as it lacks the diversity and spontaneity found in real-world multi-party interactions.

\textbf{Ubuntu Dialogue Corpus (UDC)} \cite{lowe2015ubuntu} is another widely used dataset featuring technical support conversations from the Ubuntu community. However, the dialogues in UDC are primarily noisy forum threads with fragmented turn-taking and frequent context gaps. Moreover, the conversations are largely constrained to a narrow domain of technical troubleshooting, offering limited topic diversity.

\subsection{Multilingual Capabilities of Large Language Models}
Recent multilingual large language models (MLLMs), such as GPT-4 \cite{achiam2023gpt} and LLaMA \cite{touvron2023llama}, have shown strong multilingual capabilities \cite{lai2024llms}.

Further more, Cross-lingual data sharing has been extensively studied in the context of multilingual language models (MLMs). \citet{choenni-etal-2023-languages} investigated how training samples from one language can influence the learning of other languages during fine-tuning. Similarly, \citet{doddapaneni2021primer} highlighted that multilingual models often achieve better performance by leveraging linguistic similarities across languages, enabling effective zero-shot and few-shot cross-lingual transfer.

Building upon these insights, we introduce the term \textbf{Multilingual Complementarity} to describe the phenomenon where multilingual models leverage data from multiple languages to enhance performance on a specific language task. While prior studies have observed this effect in structured tasks, its presence in complex generation tasks remains underexplored.
\section{Preliminary Experiments}
To investigate the capabilities of large language models (LLMs) in generating responses for complex dialogue scenarios, we conducted preliminary experiments using approximately 1,000 English multi-party dialogue samples. The detailed experimental setup and results are provided in Appendix \ref{appendixa}. By testing various templates and different numbers of in-context learning (ICL) examples, we observed that \textbf{LLMs failed to achieve significant improvements from ICL}. Even for English, the language LLMs are typically most proficient in, the results showed almost no variation in performance as the number of ICL examples increased.

The phenomenon raised doubts about the ability of LLMs to handle complex dialogue scenarios. What is more, \emph{if their performance on English was already so limited, would it be even worse for other languages?} To further explore this, we expanded the dataset by translating the dialogues into additional languages. Following the experimental setup in \cite{choenni-etal-2023-languages}, we tested the multilingual data under diverse templates and varying ICL example counts. The results revealed a \textbf{consistent performance decline} for all non-English languages as the number of ICL examples increased, regardless of template design. 

These findings prompt deeper questions about the nature of LLMs' capabilities and why ICL fails to yield improvements in this scenario. Much of LLMs' strength comes from the immense volume of training data they are exposed to, enabling them to approximate conversational patterns and generate contextually plausible responses. However, in complex multi-party dialogue settings, where multiple speakers interact dynamically and linguistic nuances vary significantly across languages, the models' reliance on data-driven pattern matching seems insufficient. \emph{They struggle to incorporate additional examples to refine their understanding of the dialogue's structure or adapt their reasoning to the specific task context} (\emph{\textbf{R1}}).
\section{Data Collection}
To further evaluate the capabilities of MLLMs in multi-party dialogue and to support future related research, we propose XMP dataset.
\subsection{Data Source}
We collected data from various publicly available podcasts, including "Life Kit," "Pop Culture Happy Hour," "Short Wave," "Throughline," "Consider This from NPR," "It’s Been a Minute," "The NPR Politics Podcast," "The Lawfare Podcast," and "Planet Money." All these programs are publicly accessible and available for free download. Details regarding the crawling, filtering, cleaning, and segmentation processes are provided in Appendix \ref{appendixb}. We chose podcasts as the primary data source due to their ability to provide high-quality transcripts and their frequent use of multi-party interaction formats involving three or more speakers.
\subsection{Translation Method}
\citet{singh2024translating} report that traditional machine translation systems often outperform large language models in translation quality, however, they tend to be less sensitive to cultural nuances. To validate which translation strategy is more suitable for our task, we evaluated multiple translation methods. Specifically, we experimented with commonly used tools in prior research \cite{liu2023xdailydialog,lin2021xpersona}, including Google Translate, DeepL, Baidu Translate, and iFLYTEK Translate, alongside LLaMA 3.1 and ChatGPT-4o.

We adopted both \textbf{human and automatic evaluations to ensure translation quality.} For human evaluation, we selected 50 English dialogue samples spanning diverse topics and lengths (7–15 turns), and translated them into five languages (Chinese, Japanese, French, German, Italian). Three linguists reviewed each sample for word-level accuracy, fluency, and dialogue-level coherence. \textbf{ChatGPT-4o and Google Translate received the highest overall ratings,} though ChatGPT-4o occasionally introduced off-topic phrases (e.g., “Here is the translation...”), making it less suitable for large-scale batch translation. Given its consistency,cost, speed, and prior use in similar work, we selected Google Translate.

To further support this decision, we translated 5,000 samples into Chinese and German using both GPT-4o and Google Translate, and trained LLaMA 3.1 on each dataset. The generation outcomes showed minimal differences, with Google Translate slightly outperforming in some metrics. These results reinforce its suitability for scalable multilingual dialogue construction.

Full evaluation details and translation case analyses are provided in \textbf{Appendix} \ref{appendixc}
\subsection{Dataset Quality Analysis}
As mentioned before, \citet{singh2024translating} purpose that machine translation systems tend to be less sensitive to cultural nuances. To systematically evaluate the quality of the whole dataset, we conducted a thorough assessment focusing on two key aspects: Cross-lingual Consistency Analysis and Linguistic and Cultural Consistency.
\begin{table}[!t]
	\centering
	\renewcommand\arraystretch{1.2}
	\resizebox{\linewidth}{!}{ 
		\begin{tabular}{lccc}
			\hline
			\textbf{Dataset}                  & \textbf{Task Type}                &  \textbf{Multilingual}  &\textbf{Complex Context}\\ \hline
			\textbf{UDC}   & RG              & No                      & Yes  \\ 
			\textbf{XCOPA}          & MCQA          & Yes                     &No  \\ 
			\textbf{XWinograd}          & Reasoning                             & Yes               &-   \\ 
			\textbf{XDailyDialog}          & RR                           & Yes          &No       \\ \hline
			\textbf{XMP(Ours)}          & RG                            & Yes          &Yes       \\ \hline
	\end{tabular}}
	\caption{Common multilingual or multi-party datasets. Task types include RG for Response Generation, RR for Response Retrieval, and MCQA for Multiple-Choice Question Answering.}
	\label{tab:dataset_comparison}
\end{table}
\begin{table}[!t]
	\centering
	\renewcommand\arraystretch{1.2}
	\resizebox{\linewidth}{!}{ 
		
		\begin{tabular}{lccc}
			\hline
			\textbf{Statistic}         & \textbf{Train}  & \textbf{Test}  & \textbf{Total} \\ \hline
			\#Dialogues               & 281,616         & 31,290         & 312,906        \\
			\#Utterances              & 2,384,262       & 264,888        & 2,649,150      \\
			Avg Turns per Dialogue    & 8.47            & 8.48           & 8.47           \\
			Avg Utterance Length      & 41.88           & 41.90          & 41.88          \\
			Avg Response Length       & 55.99           & 56.33          & 56.02          \\
			Avg Speakers per Dialogue & 3.63            & 3.67           & 3.64           \\
			Avg Turns per Speaker     & 2.33            & 2.31           & 2.33           \\ \hline
	\end{tabular}}
	\caption{Statistics of our dataset.}
	\label{table:dataset_statistics}
\end{table}%
Specially, we randomly selected 1000 conversations (200 samples for each language), covering various topics. For each language, two evaluation experts were selected based on their proficiency in both English and the target language. The experts were trained on the evaluation process, including specific scoring criteria and illustrative examples for each rating level, to ensure consistency and objectivity. In cases where there was significant disagreement between the two experts' scores, a third expert was consulted for adjudication. The final scores were aggregated and averaged to provide an overall quality score for each language. Detailed evaluation criteria and the scoring results for each language are provided in \textbf{Appendix} \ref{appendixd}. Overall, we found that the dataset demonstrated good quality across the board, achieving \textbf{strong results in both Cross-lingual Consistency and Linguistic and Cultural Consistency}, indicating that the translated texts can preserve semantic consistency while also taking cultural differences into account to some extent. Although relatively low-resource languages such as Italian received slightly lower scores than others, their performance still fell within an acceptable range. 

\subsection{Dataset Statistics}
In Table~\ref{tab:dataset_comparison}, we compare our XMP dataset with several commonly used multilingual or multi-party datasets. XMP is the first high-quality parallel multilingual dataset for multi-party dialogue.

\begin{figure}[!t]
	\centering
	\resizebox{0.8\linewidth}{!}{ 
		\includegraphics[width=\textwidth]{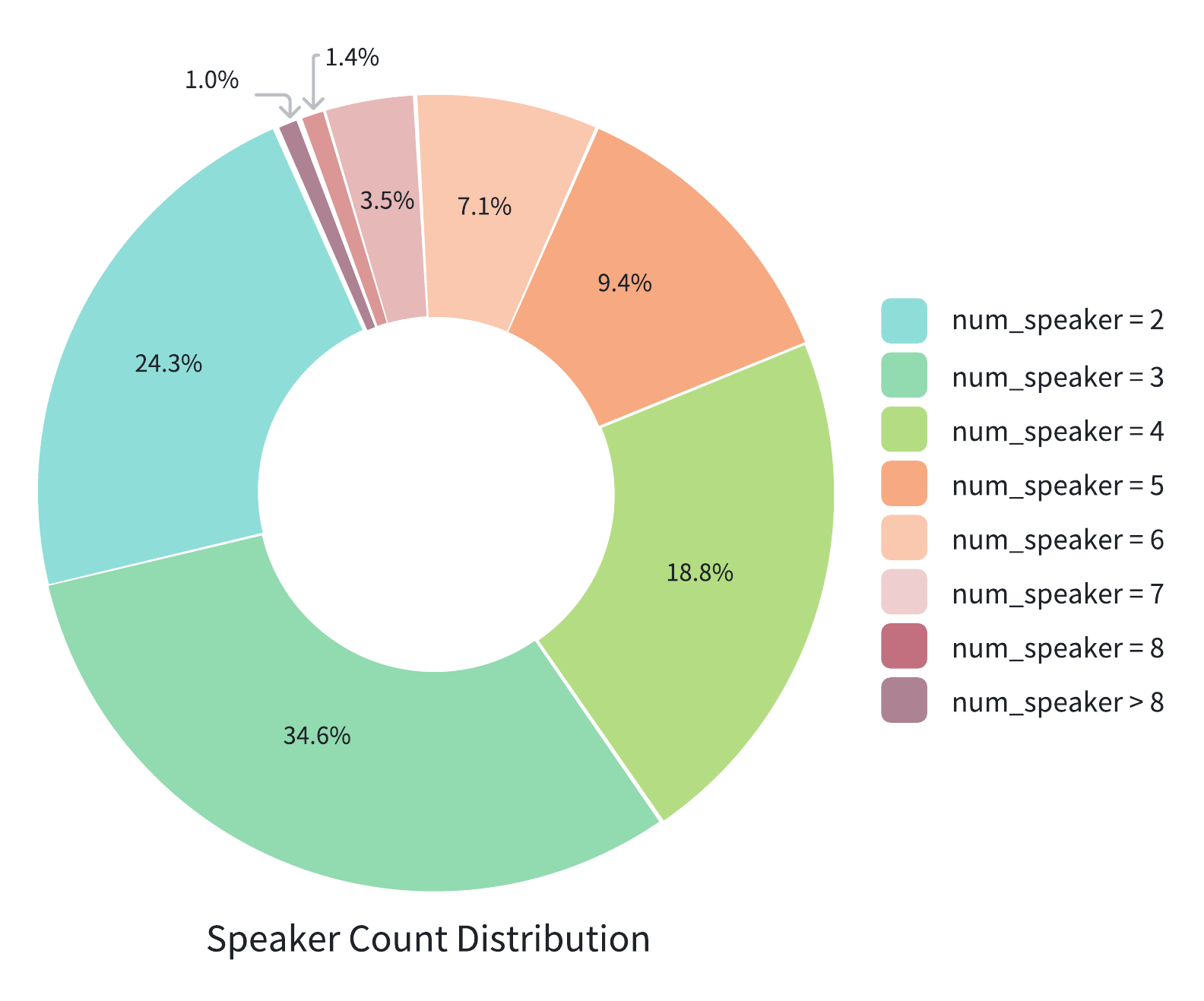}}
	\caption{Distribution of Speaker Counts}
	\label{fig:speaker_distribution}
\end{figure}

In Table \ref{table:dataset_statistics}, we present key statistics of the dataset, which includes \textbf{312k} dialogues and \textbf{2,649k} utterances, averaging \textbf{8.47} turns per dialogue. This high dialogue density supports rich multi-turn interactions. On average, each dialogue features \textbf{3.64} speakers, and over \textbf{75\%} involve at least three participants (Figure \ref{fig:speaker_distribution}), highlighting the dataset’s multi-party nature and the diverse role dynamics it captures.

Each speaker contributes an average of \textbf{2.33} turns, suggesting active participation across the dialogue. Additionally, we ensured that the \textbf{target speaker} (i.e., the speaker for whom a response is to be generated) \textbf{had spoken at least once earlier} in the dialogue history. This design guarantees that the model has sufficient contextual information to generate coherent and relevant responses.
\section{Experiments}
\subsection{Dataset Split and Baseline Models}
For training and evaluating multilingual complex dialogue scenarios, we utilized the XMP dataset, splitting it into 90\% training and 10\% testing. The dataset includes six languages: English, Chinese, Japanese, French, German, and Italian, covering a wide range of topics and multi-party interactions,  which provides a challenging benchmark for multilingual dialogue generation tasks.

To benchmark the performance of state-of-the-art large language models (LLMs) on the XMP dataset, we selected three open-source instruction-tuned models as baselines: LLaMA 3.1 8B Instruct, Qwen 2.5 7B Instruct, and LLaMA 3.1 70B Instruct. These models were chosen for their demonstrated capabilities in multilingual dialogue response task. Their inclusion allows us to examine the trade-offs between model size, and response quality across six languages.
\subsection{Task Setup}
The generation task focuses on producing high-quality responses in complex multi-party dialogue scenarios. 

Specially, Let $D={(H,S,R)}^n_{i=1}$ represents the multilingual dialogue dataset for complex scenarios, where $H$ is the dialogue history composed of multiple \emph{``Speaker: Utterance''} pairs concatenated sequentially, $S$ is the current speaker, and $R$ is the ground truth response. To guide the model in generating speaker-specific responses, a prompt is dynamically constructed based on the current speaker, such as \emph{``What will Speaker-x say next?''}. The dialogue history is then concatenated with this prompt to form the input for the model. During training, the model predicts the response $\hat{R}$, and the cross-entropy loss is computed between and the ground truth $L=-\frac{1}{m}  {\textstyle \sum_{j=1}^{m}}R_jlog\hat{R}_j $, where $m$ is the number of tokens in  $R$.
\subsection{Implementation Details}
We utilized LLaMAFactory as the training framework and applied LoRA (Low-Rank Adaptation) \cite{hu2021loralowrankadaptationlarge} for efficient fine-tuning. The training was conducted on NVIDIA A100 GPUs using the AdamW optimizer with a learning rate of $1e^{-5}$, a warm-up of 500 steps, and a batch size of 8. To enhance generalization, we standardized key entities in the dialogue history and replaced real entity data with domain-specific slot placeholders. For inference, we used vLLM, an optimized framework that ensures fast and scalable generation. Detailed training and inference commands, along with parameter settings, are provided in Appendix \ref{appendixe}.
\section{Result and Analysis}
\begin{table*}[!t]
	\centering
	\renewcommand\arraystretch{1.2}
	\resizebox{\textwidth}{!}{
	\begin{tabular}{lccccccccc}
		\toprule
		Language & Model & F1 & BLEU-1 & BLEU-2 & BLEU-3 & BLEU-4 & ROUGE-1 & ROUGE-2 & ROUGE-L \\
		\midrule
		\multirow{3}{*}{English (en)} 
		& Qwen 7B & 20.34 \textsubscript{\textcolor{red!55}{-0.89}} & 13.31 \textsubscript{\textcolor{red!30}{-0.11}} & 6.07 \textsubscript{\textcolor{red!30}{-0.04}} & 3.74 \textsubscript{\textcolor{red}{-0.12}} & 2.49 \textsubscript{\textcolor{red!30}{-0.13}} & 21.68 \textsubscript{\textcolor{red!55}{-0.91}} & 4.00 \textsubscript{\textcolor{red}{-1.60}} & 16.09 \textsubscript{\textcolor{red}{-1.65}} \\
		& LLaMA 8B & 19.92 \textsubscript{\textcolor{red}{-1.31}} & 12.80 \textsubscript{\textcolor{red!55}{-0.62}} & 5.73 \textsubscript{\textcolor{red!30}{-0.38}} & 3.45 \textsubscript{\textcolor{red!30}{-0.41}} & 2.25 \textsubscript{\textcolor{red}{-0.37}} & 21.26 \textsubscript{\textcolor{red}{-1.33}} & 3.73 \textsubscript{\textcolor{red}{-1.87}} & 15.54 \textsubscript{\textcolor{red}{-2.20}} \\
		& LLaMA 70B & 21.23 & 13.42 & 6.11 & 3.86 & 2.62 & 22.59 & 5.60 & 17.74 \\
		\midrule
		\multirow{3}{*}{Chinese (zh)} 
		& Qwen 7B & 18.65 \textsubscript{\textcolor{red}{-0.49}} & 11.45 \textsubscript{\textcolor{red!30}{-0.22}} & 4.72 \textsubscript{\textcolor{red!30}{-0.29}} & 2.78 \textsubscript{\textcolor{red}{-0.63}} & 1.77 \textsubscript{\textcolor{red!55}{-0.89}} & 19.53 \textsubscript{\textcolor{red!55}{-0.84}} & 2.60 \textsubscript{\textcolor{red}{-1.45}} & 15.74 \textsubscript{\textcolor{red}{-1.15}} \\
		& LLaMA 8B & 18.37 \textsubscript{\textcolor{red!55}{-0.77}} & 11.60 \textsubscript{\textcolor{teal}{+0.23}} & 4.68 \textsubscript{\textcolor{red!30}{-0.33}} & 2.76 \textsubscript{\textcolor{red!55}{-0.65}} & 1.76 \textsubscript{\textcolor{red!55}{-0.90}} & 19.26 \textsubscript{\textcolor{red}{-1.11}} & 2.45 \textsubscript{\textcolor{red}{-1.60}} & 15.44 \textsubscript{\textcolor{red}{-1.45}} \\
		& LLaMA 70B & 19.14 & 11.37 & 5.01 & 3.41 & 2.66 & 20.37 & 4.05 & 16.89 \\
		\midrule
		\multirow{3}{*}{Japanese (ja)} 
		& Qwen 7B & 28.17 \textsubscript{\textcolor{teal}{+1.07}} & 17.74 \textsubscript{\textcolor{teal}{+1.16}} & 9.27 \textsubscript{\textcolor{teal}{+0.31}} & 5.98 \textsubscript{\textcolor{teal}{+0.59}} & 3.96 \textsubscript{\textcolor{teal}{+0.19}} & 34.76 \textsubscript{\textcolor{teal}{+1.81}} & 9.16 \textsubscript{\textcolor{teal}{+0.32}} & 19.08 \textsubscript{\textcolor{red!55}{-0.99}} \\
		& LLaMA 8B & 27.69 \textsubscript{\textcolor{teal}{+0.59}} & 16.81 \textsubscript{\textcolor{teal}{+0.23}} & 8.76 \textsubscript{\textcolor{red!30}{-0.20}} & 5.72 \textsubscript{\textcolor{red!55}{-0.67}} & 3.90 \textsubscript{\textcolor{red}{-0.87}} & 33.71 \textsubscript{\textcolor{teal}{+0.76}} & 8.73 \textsubscript{\textcolor{red}{-1.11}} & 19.08 \\
		& LLaMA 70B & 27.10 & 15.58 & 8.96 & 6.39 & 4.77 & 32.95 & 9.84 & 20.07 \\
		\midrule
		\multirow{3}{*}{French (fr)} 
		& Qwen 7B & 16.35 \textsubscript{\textcolor{red}{-1.13}} & 10.14 \textsubscript{\textcolor{red!30}{-0.27}} & 4.31 \textsubscript{\textcolor{red!30}{-0.09}} & 2.59 \textsubscript{\textcolor{red!30}{-0.25}} & 1.68 \textsubscript{\textcolor{red!30}{-0.13}} & 17.44 \textsubscript{\textcolor{red}{-1.25}} & 2.41 \textsubscript{\textcolor{red}{-1.63}} & 13.20 \textsubscript{\textcolor{red}{-1.78}} \\
		& LLaMA 8B & 17.20 \textsubscript{\textcolor{red!30}{-0.28}} & 11.21 \textsubscript{\textcolor{teal}{+0.80}} & 4.87 \textsubscript{\textcolor{teal}{+0.47}} & 3.00 \textsubscript{\textcolor{teal}{+0.16}} & 2.00 \textsubscript{\textcolor{teal}{+0.19}} & 18.34 \textsubscript{\textcolor{red!30}{-0.35}} & 2.85 \textsubscript{\textcolor{red}{-1.19}} & 13.63 \textsubscript{\textcolor{red}{-1.35}} \\
		& LLaMA 70B & 17.48 & 10.41 & 4.40 & 2.84 & 1.81 & 18.69 & 4.04 & 14.98 \\
		\midrule
		\multirow{3}{*}{German (de)} 
		& Qwen 7B & 17.36 \textsubscript{\textcolor{red}{-1.09}} & 11.06 \textsubscript{\textcolor{red!30}{-0.34}} & 4.68 \textsubscript{\textcolor{red}{-1.03}} & 2.76 \textsubscript{\textcolor{red}{-1.19}} & 1.77 \textsubscript{\textcolor{red}{-1.22}} & 17.89 \textsubscript{\textcolor{red}{-1.23}} & 2.55 \textsubscript{\textcolor{red}{-1.48}} & 14.21 \textsubscript{\textcolor{red}{-1.72}} \\
		& LLaMA 8B & 18.09 \textsubscript{\textcolor{red!30}{-0.36}} & 12.05 \textsubscript{\textcolor{teal}{+0.65}} & 5.18 \textsubscript{\textcolor{red!55}{-0.53}} & 3.07 \textsubscript{\textcolor{red!55}{-0.88}} & 2.01 \textsubscript{\textcolor{red!55}{-0.98}} & 18.54 \textsubscript{\textcolor{red!55}{-0.58}} & 2.86 \textsubscript{\textcolor{red}{-1.17}} & 14.53 \textsubscript{\textcolor{red}{-1.40}} \\
		& LLaMA 70B & 18.45 & 11.40 & 5.71 & 3.95 & 2.99 & 19.12 & 4.03 & 15.93 \\
		\midrule
		\multirow{3}{*}{Italian (it)} 
		& Qwen 7B & 16.42 \textsubscript{\textcolor{red}{-1.01}} & 10.26 \textsubscript{\textcolor{red!30}{-0.41}} & 4.20 \textsubscript{\textcolor{red}{-1.27}} & 2.49 \textsubscript{\textcolor{red}{-1.40}} & 1.59 \textsubscript{\textcolor{red}{-1.40}} & 17.03 \textsubscript{\textcolor{red}{-1.13}} & 2.09 \textsubscript{\textcolor{red}{-1.75}} & 13.38 \textsubscript{\textcolor{red}{-1.64}} \\
		& LLaMA 8B & 17.12 \textsubscript{\textcolor{red!30}{-0.31}} & 11.01 \textsubscript{\textcolor{teal}{+0.34}} & 4.65 \textsubscript{\textcolor{red!55}{-0.82}} & 2.81 \textsubscript{\textcolor{red}{-1.08}} & 1.83 \textsubscript{\textcolor{red}{-1.16}} & 17.70 \textsubscript{\textcolor{red!30}{-0.46}} & 2.48 \textsubscript{\textcolor{red}{-1.36}} & 13.79 \textsubscript{\textcolor{red}{-1.23}} \\
		& LLaMA 70B & 17.43 & 10.67 & 5.47 & 3.89 & 2.99 & 18.16 & 3.84 & 15.02 \\
		\bottomrule	\end{tabular}}
	\caption{Performance comparison of Qwen-2.5 7B, LLaMA-3.1 8B, and LLaMA-3.1 70B across different languages. The differences between Qwen-2.5 7B and LLaMA-3.1 8B relative to LLaMA-3.1 70B for each metric are calculated and highlighted.}
	\label{tab:results}
\end{table*}
\subsection{Model Performance}
As shown in Table \ref{tab:results}, we evaluate Qwen-2.5 7B, LLaMA-3.1 8B, and LLaMA-3.1 70B on the XMP dataset across six languages: English, Chinese, Japanese, French, German, and Italian. For text tokenization, Jieba was used for Chinese, Janome for Japanese, and NLTK for the remaining languages. 

English outperforms other languages on most metrics, except for Japanese. The result aligns with expectations, as English occupies a significant proportion of the pretraining data for these models. The results highlight the over-reliance of LLMs on high-resource languages.

\emph{The performance gain of LLaMA-3.1 70B over smaller models (e.g., 8B or Qwen-2.5 7B) is marginal, with improvements typically within 1\% across most metrics (\textbf{R2})}. This suggests that the emergent abilities often associated with larger models are not prominently reflected in this task. Instead, the slight advantage may stem from the larger models' capacity to generate grammatically and lexically refined text, rather than demonstrating a better understanding of the task.
\subsection{Structural and Semantic divergence}
Japanese and Chinese both achieve relatively high scores on surface-level metrics like BLEU and ROUGE, Japanese even outperforming all other languages besides English, and Chinese ranking among the top as well. This can be attributed to factors such as syntactic regularity, frequent use of fixed expressions and kana in Japanese, and strong n-gram overlaps that benefit token-based evaluation.

However, as shown in Table \ref{tab:bertscore_results}, BERTScore tells a different story: Japanese aligns semantically with other non-English languages, while Chinese records the lowest score (0.6086), indicating significant structural and semantic divergence that models struggle to capture.
\begin{table}[!t]
	\centering
	\renewcommand\arraystretch{1.2}
	\resizebox{0.5\linewidth}{!}{
	\begin{tabular}{cc}
		\hline
		\textbf{Language} & \textbf{BERTScore} \\
		\hline
		EN & 0.8507 \\
		ZH & 0.6086 \\
		JA & 0.6730 \\
		FR & 0.6707 \\
		DE & 0.6640 \\
		IT & 0.6657 \\
		\hline
	\end{tabular}}
	\caption{BERTScore results for LLaMA-3.1 8B across different languages.}
	\label{tab:bertscore_results}
\end{table}
\subsection{Multilingual Complementary Ability}
\begin{table*}[h!]
	\centering
	\renewcommand\arraystretch{1.2}
	\resizebox{\linewidth}{!}{
	\begin{tabular}{lccccccccc}
		\hline
		 \textbf{Train Data} & \textbf{Test Data} & \textbf{F1} & \textbf{BLEU-1} & \textbf{BLEU-2} & \textbf{BLEU-3} & \textbf{BLEU-4} & \textbf{ROUGE-1} &\textbf{ROUGE-2} &\textbf{ROUGE-L} \\ \hline
		\multicolumn{10}{c}{\textbf{LLaMA-8B}}                                                                                                   \\ \hline
		DE (single)             & DE                     & 18.09                    & 12.05                      & 5.18                      & 3.07                      & 2.01                      & 18.54                      & 2.86                      & 14.53                      \\ 
		DE, FR (mixed)          & DE                     & 16.58\textsubscript{\textcolor{red}{-1.51}}  & 9.87\textsubscript{\textcolor{red}{-2.18}}  & 4.25\textsubscript{\textcolor{red}{-0.93}} & 2.64\textsubscript{\textcolor{red!50}{-0.43}} & 1.79\textsubscript{\textcolor{red!50}{-0.22}} & 17.34\textsubscript{\textcolor{red}{-1.20}} & 2.40\textsubscript{\textcolor{red!50}{-0.46}} & 14.06\textsubscript{\textcolor{red!50}{-0.47}} \\ 
		DE, ZH (mixed)          & DE                     & 16.49\textsubscript{\textcolor{red}{-1.60}}  & 9.72\textsubscript{\textcolor{red}{-2.33}}  & 4.15\textsubscript{\textcolor{red}{-1.03}} & 2.56\textsubscript{\textcolor{red}{-0.51}} & 1.72\textsubscript{\textcolor{red!50}{-0.29}} & 17.28\textsubscript{\textcolor{red}{-1.26}} & 2.36\textsubscript{\textcolor{red}{-0.50}} & 13.94\textsubscript{\textcolor{red}{-0.59}} \\ 
		EN, DE (mixed)          & DE                     & 16.48\textsubscript{\textcolor{red}{-1.61}}  & 9.86\textsubscript{\textcolor{red}{-2.19}}  & 4.14\textsubscript{\textcolor{red}{-1.04}} & 2.52\textsubscript{\textcolor{red}{-0.55}} & 1.66\textsubscript{\textcolor{red!50}{-0.35}} & 17.24\textsubscript{\textcolor{red}{-1.30}} & 2.23\textsubscript{\textcolor{red}{-0.63}} & 13.94\textsubscript{\textcolor{red}{-0.59}} \\ 
		 
		 \hline
		FR (single)             & FR                     & 17.20                    & 11.21                     & 4.87                      & 3.00                      & 2.00                      & 18.34                      & 2.85                      & 13.63                      \\ 
		DE, FR (mixed)          & FR                     & 15.66\textsubscript{\textcolor{red}{-1.54}}  & 9.15\textsubscript{\textcolor{red}{-2.06}}  & 3.99\textsubscript{\textcolor{red}{-0.88}}  & 2.53\textsubscript{\textcolor{red!50}{-0.47}}  & 1.71\textsubscript{\textcolor{red!50}{-0.29}}  & 16.79\textsubscript{\textcolor{red}{-1.55}}  & 2.34\textsubscript{\textcolor{red}{-0.51}}  & 13.19\textsubscript{\textcolor{red!50}{-0.44}} \\ 
		EN, FR (mixed)          & FR                     & 15.59\textsubscript{\textcolor{red}{-1.61}}  & 9.02\textsubscript{\textcolor{red}{-2.19}}  & 3.94\textsubscript{\textcolor{red}{-0.93}}  & 2.48\textsubscript{\textcolor{red}{-0.52}}  & 1.66\textsubscript{\textcolor{red}{-0.34}}  & 16.73\textsubscript{\textcolor{red}{-1.61}}  & 2.33\textsubscript{\textcolor{red}{-0.52}}  & 13.10\textsubscript{\textcolor{red}{-0.53}} \\ 
		 
		 \hline
		ZH (single)             & ZH                     & 18.37                    & 11.60                     & 4.68                      & 2.76                      & 1.76                      & 19.26                      & 2.45                      & 15.44                      \\ 
		DE, ZH (mixed)          & ZH                     & 17.70\textsubscript{\textcolor{red}{-0.67}}  & 10.24\textsubscript{\textcolor{red}{-1.36}}  & 4.29\textsubscript{\textcolor{red!50}{-0.39}}  & 2.65\textsubscript{\textcolor{red!50}{-0.11}}  & 1.76\textsubscript{\textcolor{teal!50}{+0.00}}  & 18.89\textsubscript{\textcolor{red!50}{-0.37}}  & 2.43\textsubscript{\textcolor{red!50}{-0.02}}  & 15.45\textsubscript{\textcolor{teal!50}{+0.01}} \\ 
		EN, ZH (mixed)          & ZH                     & 17.69\textsubscript{\textcolor{red}{-0.68}}  & 10.18\textsubscript{\textcolor{red}{-1.42}}  & 4.23\textsubscript{\textcolor{red}{-0.45}}  & 2.57\textsubscript{\textcolor{red!50}{-0.19}}  & 1.68\textsubscript{\textcolor{red!50}{-0.08}}  & 18.90\textsubscript{\textcolor{red!50}{-0.36}}  & 2.37\textsubscript{\textcolor{red!50}{-0.08}}  & 15.36\textsubscript{\textcolor{red!50}{-0.08}} \\ 
		 
		 \hline
		EN (single)             & EN                     & 19.92                    & 12.80                     & 5.73                      & 3.45                      & 2.25                      & 21.26                      & 3.73                      & 15.54                      \\ 
		EN, DE (mixed)          & EN                     & 19.73\textsubscript{\textcolor{red!50}{-0.19}}  & 12.25\textsubscript{\textcolor{red!50}{-0.55}}  & 5.79\textsubscript{\textcolor{teal!50}{+0.06}}  & 3.73\textsubscript{\textcolor{teal!50}{+0.28}}  & 2.62\textsubscript{\textcolor{teal!50}{+0.37}}  & 21.17\textsubscript{\textcolor{red!50}{-0.09}}  & 4.08\textsubscript{\textcolor{teal!50}{+0.35}}  & 16.13\textsubscript{\textcolor{teal}{+0.59}} \\ 
		EN, FR (mixed)          & EN                     & 19.74\textsubscript{\textcolor{red!50}{-0.18}}  & 12.26\textsubscript{\textcolor{red}{-0.54}}  & 5.77\textsubscript{\textcolor{teal!50}{+0.04}}  & 3.68\textsubscript{\textcolor{teal!50}{+0.23}}  & 2.54\textsubscript{\textcolor{teal!50}{+0.29}}  & 21.18\textsubscript{\textcolor{red!50}{-0.08}}  & 4.10\textsubscript{\textcolor{teal!50}{+0.37}}  & 16.14\textsubscript{\textcolor{teal}{+0.60}} \\ 
		EN, ZH (mixed)          & EN                     & 19.70\textsubscript{\textcolor{red!50}{-0.22}}  & 12.18\textsubscript{\textcolor{red}{-0.62}}  & 5.74\textsubscript{\textcolor{teal!50}{+0.01}}  & 3.70\textsubscript{\textcolor{teal!50}{+0.25}}  & 2.57\textsubscript{\textcolor{teal!50}{+0.32}}  & 21.27\textsubscript{\textcolor{teal!50}{+0.01}}  & 4.10\textsubscript{\textcolor{teal!50}{+0.37}}  & 16.12\textsubscript{\textcolor{teal}{+0.58}} \\ 
		 
		 \hline
		\multicolumn{10}{c}{\textbf{LLaMA-70B}}                                                                                                  \\ \hline
		DE (single)             & DE                     & 18.45                    & 11.40                     & 5.71                      & 3.95                      & 2.99                      & 19.12                      & 4.03                      & 15.93                      \\ 
		DE, FR (mixed)          & DE                     & 18.45\textsubscript{\textcolor{red!50}{-0.00}}  & 11.44\textsubscript{\textcolor{teal!50}{+0.04}}  & 5.70\textsubscript{\textcolor{red!50}{-0.01}}  & 3.85\textsubscript{\textcolor{red!50}{-0.10}}  & 3.05\textsubscript{\textcolor{teal!50}{+0.06}}  & 19.58\textsubscript{\textcolor{teal!50}{+0.46}}  & 4.30\textsubscript{\textcolor{teal!50}{+0.27}}  & 16.23\textsubscript{\textcolor{teal!50}{+0.30}} \\ 
		DE, ZH (mixed)          & DE                     & 18.49\textsubscript{\textcolor{teal!50}{+0.04}}  & 11.37\textsubscript{\textcolor{red!50}{-0.03}}  & 5.80\textsubscript{\textcolor{teal!50}{+0.09}}  & 3.91\textsubscript{\textcolor{red}{-0.04}}  & 3.06\textsubscript{\textcolor{teal!50}{+0.07}}  & 19.01\textsubscript{\textcolor{red}{-0.11}}  & 4.15\textsubscript{\textcolor{teal!50}{+0.12}}  & 15.88\textsubscript{\textcolor{red!50}{-0.05}} \\ 
		EN, DE (mixed)          & DE                     & 17.00\textsubscript{\textcolor{red}{-1.45}}  & 11.70\textsubscript{\textcolor{teal!50}{+0.30}}  & 5.12\textsubscript{\textcolor{red}{-0.59}}  & 3.08\textsubscript{\textcolor{red}{-0.87}}  & 2.06\textsubscript{\textcolor{red}{-0.93}}  & 17.79\textsubscript{\textcolor{red}{-1.33}}  & 2.97\textsubscript{\textcolor{red}{-1.06}}  & 13.30\textsubscript{\textcolor{red}{-2.63}} \\

		 \hline
		FR (single)             & FR                     & 17.48                    & 10.41                     & 5.40                      & 3.84                      & 2.96                      & 18.69                      & 4.04                      & 14.98                      \\ 
		DE, FR (mixed)          & FR                     & 18.14\textsubscript{\textcolor{teal}{+0.66}}  & 11.20\textsubscript{\textcolor{teal}{+0.79}}  & 5.88\textsubscript{\textcolor{teal!50}{+0.48}}  & 4.16\textsubscript{\textcolor{teal!50}{+0.32}}  & 3.17\textsubscript{\textcolor{teal!50}{+0.21}}  & 19.26\textsubscript{\textcolor{teal}{+0.57}}  & 4.45\textsubscript{\textcolor{teal!50}{+0.41}}  & 15.40\textsubscript{\textcolor{teal!50}{+0.42}} \\ 
		EN, FR (mixed)          & FR                     & 18.59\textsubscript{\textcolor{teal}{+1.11}}  & 12.14\textsubscript{\textcolor{teal}{+1.73}}  & 5.98\textsubscript{\textcolor{teal}{+0.58}}  & 4.04\textsubscript{\textcolor{teal!50}{+0.20}}  & 2.99\textsubscript{\textcolor{teal!50}{+0.03}}  & 19.42\textsubscript{\textcolor{teal}{+0.73}}  & 4.11\textsubscript{\textcolor{teal!50}{+0.07}}  & 15.18\textsubscript{\textcolor{teal!50}{+0.20}} \\ 
		 
		\hline
		ZH (single)             & ZH                     & 19.14                    & 11.37                     & 5.70                      & 4.01                      & 3.06                      & 20.37                      & 4.05                      & 16.89                      \\ 
		DE, ZH (mixed)          & ZH                     & 19.06\textsubscript{\textcolor{red!50}{-0.08}}  & 11.33\textsubscript{\textcolor{red!50}{-0.04}}  & 5.74\textsubscript{\textcolor{teal!50}{+0.04}}  & 4.10\textsubscript{\textcolor{teal!50}{+0.09}}  & 3.12\textsubscript{\textcolor{teal!50}{+0.06}}  & 20.33\textsubscript{\textcolor{red!50}{-0.04}}  & 4.25\textsubscript{\textcolor{teal!50}{+0.20}}  & 17.17\textsubscript{\textcolor{teal!50}{+0.28}} \\  
		EN, ZH (mixed)          & ZH                     & 19.08\textsubscript{\textcolor{red!50}{-0.06}}  & 11.26\textsubscript{\textcolor{red!50}{-0.11}}  & 5.53\textsubscript{\textcolor{red!50}{-0.17}}  & 4.01\textsubscript{\textcolor{red!50}{-0.00}}  & 3.03\textsubscript{\textcolor{red!50}{-0.03}}  & 20.30\textsubscript{\textcolor{red!50}{-0.07}}  & 4.13\textsubscript{\textcolor{teal!50}{+0.08}}  & 17.07\textsubscript{\textcolor{teal!50}{+0.18}} \\

		\hline
		EN (single)             & EN                     & 21.23                    & 13.42                     & 7.11                      & 4.98                      & 3.76                      & 22.59                      & 5.60                      & 17.74                      \\ 
		EN, DE (mixed)          & EN                     & 17.83\textsubscript{\textcolor{red}{-3.40}}  & 11.87\textsubscript{\textcolor{red}{-1.55}}  & 4.90\textsubscript{\textcolor{red}{-2.21}}  & 2.81\textsubscript{\textcolor{red}{-2.17}}  & 1.76\textsubscript{\textcolor{red}{-2.00}}  & 19.37\textsubscript{\textcolor{red}{-3.22}}  & 2.77\textsubscript{\textcolor{red}{-2.83}}  & 13.10\textsubscript{\textcolor{red}{-4.64}} \\ 
		EN, FR (mixed)          & EN                     & 21.94\textsubscript{\textcolor{teal}{+0.71}}  & 14.52\textsubscript{\textcolor{teal}{+1.10}}  & 7.69\textsubscript{\textcolor{teal}{+0.58}}  & 5.40\textsubscript{\textcolor{teal!50}{+0.42}}  & 4.10\textsubscript{\textcolor{teal!50}{+0.34}}  & 23.17\textsubscript{\textcolor{teal}{+0.58}}  & 5.91\textsubscript{\textcolor{teal!50}{+0.31}}  & 17.98\textsubscript{\textcolor{teal!50}{+0.24}} \\ 
		EN, ZH (mixed)          & EN                     & 21.67\textsubscript{\textcolor{teal!50}{+0.44}}  & 13.51\textsubscript{\textcolor{teal!50}{+0.09}}  & 7.18\textsubscript{\textcolor{teal!50}{+0.07}}  & 4.96\textsubscript{\textcolor{red!50}{-0.02}}  & 3.72\textsubscript{\textcolor{red!50}{-0.04}}  & 23.27\textsubscript{\textcolor{teal}{+0.68}}  & 5.61\textsubscript{\textcolor{teal!50}{+0.01}}  & 17.76\textsubscript{\textcolor{teal!50}{+0.02}} \\ 
		\hline
	\end{tabular}}
	\caption{Comparison of multilingual complementary ability for LLaMA-3.1 8B and LLaMA-3.1 70B, grouped by testing language. Results include single- and mixed-language training scenarios.}
	\label{tab:multilingual_complementary_grouped}
\end{table*}
Multilingual complementary ability is defined as a model's capacity to effectively leverage information from multiple languages during training to enhance performance on specific target languages. The results of our experiments, as detailed in Table \ref{tab:multilingual_complementary_grouped}, shed light on this phenomenon by using LLaMA-3.1 8B and LLaMA-3.1 70B models tested on the XMP dataset. Notably, our findings diverge significantly from conventional understanding in prior research.
\subsubsection{Mixed Training with English}
Previous researches \cite{whitehouse2023llm,li-etal-2024-prealign} often highlight that mixing English as a source and target language during training provides significant benefits for the target language due to English's dominance in pretraining corpora and its rich semantic structure. However, our experiments on XMP challenge this assumption. For the LLaMA-3.1 8B model, mixing English with any other language consistently leads to negative performance impacts. Interestingly, for the LLaMA-3.1 70B model, this detrimental effect is mitigated, particularly when English is mixed with French. However, even in cases where certain languages exhibit improvements in specific metrics, these gains remain minimal, typically less than 0.5\%. This suggests that while larger model capacities can temper the negative impacts of mixed training, \emph{the benefits derived from cross-lingual interaction are marginal and insufficient to fully leverage the potential of multilingual complementary ability in complex dialogue scenarios. \textbf{(R3)}}
\subsubsection{Cross-Language Benefits Are Asymmetrical}
The experimental results for 70B-model reveal an intriguing asymmetry: while English typically fails to enhance the performance of other languages, other languages, such as French or German, often provide improvements when mixed with English. 
\subsubsection{Effect of Linguistic Similarity}
Both \citet{eronen2023zero} and \citet{aepli2022improving} report that linguistic similarity plays a crucial role in cross-lingual transfer. \citet{eronen2023zero} demonstrate this across sentiment analysis, and dependency parsing. Similarly, \citet{aepli2022improving} observe that in POS tagging, transfer between closely related language varieties yields significantly better performance than between unrelated pairs.

However, our experimental findings suggest that these linguistic connections do not guarantee consistent or significant cross-lingual improvements. For instance, the mixing of English and German often results in negative impacts on both languages' performance, even though they are closely related in terms of linguistic roots. Conversely, the pairing of French with either German or English generally produces positive outcomes, albeit with varying degrees of improvement depending on the specific metrics and target languages. 

The nuanced relationship highlights the need for deeper investigations into how linguistic similarity interacts with other variables in multilingual training, particularly in complex and dynamic tasks. It underscores that shared linguistic heritage, while beneficial, is not a deterministic factor for cross-lingual success, and the benefits of multilingual training are far from uniform across different language pairs.
\subsection{Probing Context Sensitivity}
\begin{figure*}[!t]
	\centering
	\resizebox{0.9\linewidth}{!}{ 
		\includegraphics[width=\textwidth]{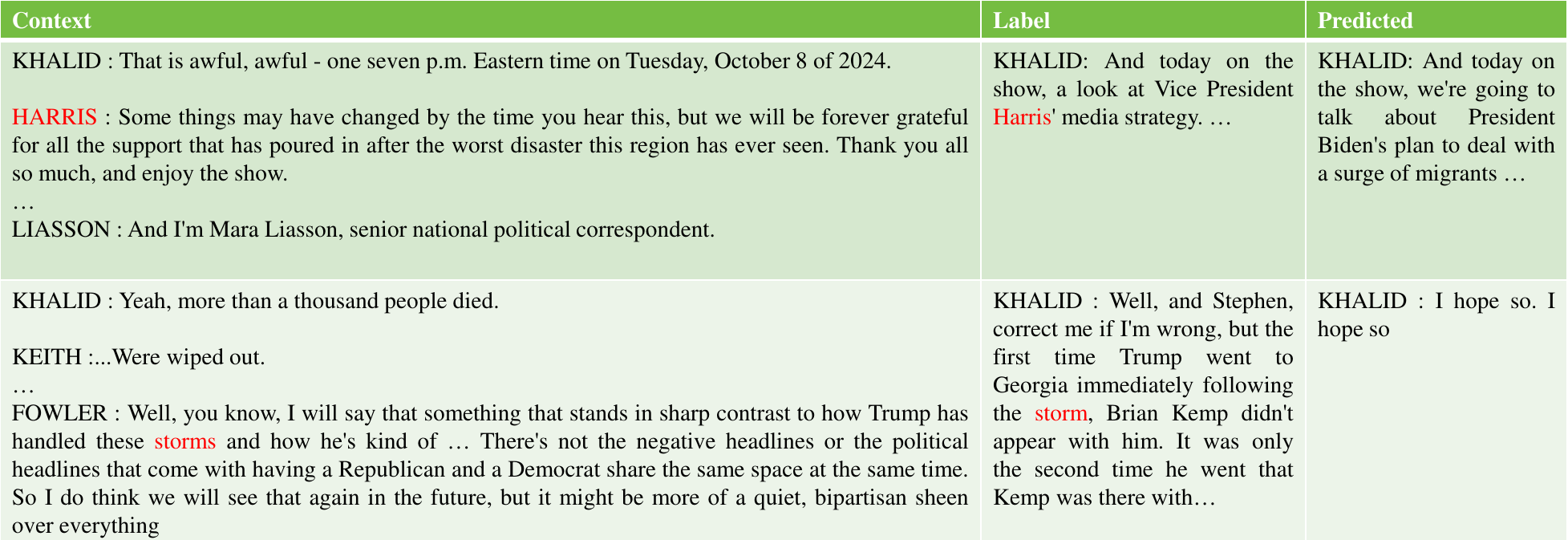}}
	\caption{Case Study}
	\label{fig:case}
\end{figure*}
To examine how effectively the model leverages extended dialogue history, we conducted a controlled evaluation by randomly removing 30\% of the context while retaining the final two turns. The rationale behind this setup is that if the model truly integrates long-range contextual information, removing part of the earlier context should noticeably impact its performance.

We applied this evaluation to the LLaMA 3 8B model on both English and Chinese subsets. Results are presented in Table~\ref{tab:context_removal}. Overall, the differences across both languages were minimal.

These findings suggest that the model does not effectively utilize broader dialogue history. Its performance appears to rely primarily on recent turns, reflecting a limited capacity to model longer-range conversational dependencies.
\begin{table}[t!]
	\centering
	\resizebox{\linewidth}{!}{
	\begin{tabular}{llcccc}
		\toprule
		\textbf{Language} & \textbf{Context} & \textbf{F1} & \textbf{B-1} & \textbf{B-2} & \textbf{R-L} \\
		\midrule
		Chinese & Full                 & 18.37 & 11.60 & 4.68 & 5.44 \\
		Chinese & - 30\% & 18.52 & 12.19 & 4.77 & 5.85 \\
		English & Full                & 19.92 & 12.80 & 5.73 & 5.54 \\
		English & - 30\% & 19.36 & 12.60 & 5.41 & 4.91 \\
		\bottomrule
	\end{tabular}}
	\caption{Performance comparison with and without 30\% context removal on the LLaMA 3 8B model. Context removal excludes the final two turns.}
	\label{tab:context_removal}
\end{table}

\subsection{Case Study}
To more intuitively demonstrate the current limitations of large language models in understanding multi-party conversations, we conduct a case study and analyze the model's predictions. We found that the model's responses often: 1. Misinterpreted conversational context and overlooked the roles of specific participants. 2. Lacked concrete factual content

We presented several cases in the Figure \ref{fig:case} . Due to character limitations, all examples shown here are in English — the model’s performance tends to degrade further in other languages.

In the first case, the model ignored the speaker Harris and abruptly shifted to a discussion of \emph{Biden’s plan}. In contrast, the ground truth continued with a discussion of Harris’ media strategy, which was contextually appropriate. We attribute this error to amplified hallucination under complex dialogue settings.

In the second example, the model predicted a response like “I hope so,” which contributed nothing meaningful to the conversation, whereas the ground truth was a natural continuation of the ongoing topic. We believe this failure is likely due to: 1. the model struggles to maintain awareness of the broader conversational flow and context, particularly in multi-party settings; 2. current large models' pretraining corpora still lack rich, multi-turn, real-world conversational data.
\section{Conclusion}
This study aims to explore the generalization capabilities of MLLM in complex dialogue scenarios. To enable this investigation, we construct \textbf{XMP}, a high-quality, parallel multilingual multi-party dialogue dataset. XMP serves as a critical resource to support rigorous evaluation in this challenging setting.

Through extensive experiments, we find:
\textbf{R1}: MLLMs exhibit significant limitations in generalizing to multi-party dialogue.
\textbf{R2}: Fine-tuning on XMP yields only marginal improvements, with larger models showing limited gains.
\textbf{R3}: Contrary to prior studies, multilingual complementarity effects are inconsistent or even detrimental.

We hope our work and the XMP dataset will inspire further efforts toward building more context-aware, robust, and multilingual conversational AI.
\section{Limitations}
Despite the contributions of this study, several limitations warrant discussion. First, the XMP dataset currently includes a limited number of languages, omitting many low-resource languages, which constrains its generalizability to broader multilingual contexts. Second, during the dataset collection process, the team also collected corresponding podcast audio data. However, the significant challenges associated with audio preprocessing have so far prevented XMP from being expanded into a multimodal corpus. This limitation is a priority for future improvement, with plans to incorporate audio features to enhance the dataset's utility. Finally, the exploration of linguistic similarity in this study was limited to a few language pairs, primarily focusing on German, French, English, and Chinese, leaving the effects of cross-lingual interactions among other linguistic families underexplored. Addressing these limitations could significantly enhance the applicability and impact of XMP in future multilingual and multimodal research.
\bibliography{custom}

\begin{thebibliography}{25}
\providecommand{\natexlab}[1]{#1}

\bibitem[{Achiam et~al.(2023)Achiam, Adler, Agarwal, Ahmad, Akkaya, Aleman,
  Almeida, Altenschmidt, Altman, Anadkat et~al.}]{achiam2023gpt}
Josh Achiam, Steven Adler, Sandhini Agarwal, Lama Ahmad, Ilge Akkaya,
  Florencia~Leoni Aleman, Diogo Almeida, Janko Altenschmidt, Sam Altman,
  Shyamal Anadkat, et~al. 2023.
\newblock Gpt-4 technical report.
\newblock \emph{arXiv preprint arXiv:2303.08774}.

\bibitem[{Aepli and Sennrich(2022)}]{aepli2022improving}
No{\"e}mi Aepli and Rico Sennrich. 2022.
\newblock Improving zero-shot cross-lingual transfer between closely related
  languages by injecting character-level noise.
\newblock In \emph{Findings of the Association for Computational Linguistics:
  ACL 2022}, pages 4074--4083.

\bibitem[{Choenni et~al.(2023)Choenni, Garrette, and
  Shutova}]{choenni-etal-2023-languages}
Rochelle Choenni, Dan Garrette, and Ekaterina Shutova. 2023.
\newblock \href {https://doi.org/10.18653/v1/2023.emnlp-main.818} {How do
  languages influence each other? studying cross-lingual data sharing during
  {LM} fine-tuning}.
\newblock In \emph{Proceedings of the 2023 Conference on Empirical Methods in
  Natural Language Processing}, pages 13244--13257, Singapore. Association for
  Computational Linguistics.

\bibitem[{Conneau et~al.(2020{\natexlab{a}})Conneau, Khandelwal, Goyal,
  Chaudhary, Wenzek, Guzm{\'a}n, Grave, Ott, Zettlemoyer, and
  Stoyanov}]{conneau-etal-2020-unsupervised}
Alexis Conneau, Kartikay Khandelwal, Naman Goyal, Vishrav Chaudhary, Guillaume
  Wenzek, Francisco Guzm{\'a}n, Edouard Grave, Myle Ott, Luke Zettlemoyer, and
  Veselin Stoyanov. 2020{\natexlab{a}}.
\newblock \href {https://doi.org/10.18653/v1/2020.acl-main.747} {Unsupervised
  cross-lingual representation learning at scale}.
\newblock In \emph{Proceedings of the 58th Annual Meeting of the Association
  for Computational Linguistics}, pages 8440--8451, Online. Association for
  Computational Linguistics.

\bibitem[{Conneau et~al.(2020{\natexlab{b}})Conneau, Wu, Li, Zettlemoyer, and
  Stoyanov}]{conneau2020emerging}
Alexis Conneau, Shijie Wu, Haoran Li, Luke Zettlemoyer, and Veselin Stoyanov.
  2020{\natexlab{b}}.
\newblock Emerging cross-lingual structure in pretrained language models.
\newblock In \emph{Proceedings of the 58th Annual Meeting of the Association
  for Computational Linguistics}, pages 6022--6034.

\bibitem[{Doddapaneni et~al.(2021)Doddapaneni, Ramesh, Khapra, Kunchukuttan,
  and Kumar}]{doddapaneni2021primer}
Sumanth Doddapaneni, Gowtham Ramesh, Mitesh~M Khapra, Anoop Kunchukuttan, and
  Pratyush Kumar. 2021.
\newblock A primer on pretrained multilingual language models.
\newblock \emph{arXiv preprint arXiv:2107.00676}.

\bibitem[{Eronen et~al.(2023)Eronen, Ptaszynski, and Masui}]{eronen2023zero}
Juuso Eronen, Michal Ptaszynski, and Fumito Masui. 2023.
\newblock Zero-shot cross-lingual transfer language selection using linguistic
  similarity.
\newblock \emph{Information Processing \& Management}, 60(3):103250.

\bibitem[{Hu et~al.(2021)Hu, Shen, Wallis, Allen-Zhu, Li, Wang, Wang, and
  Chen}]{hu2021loralowrankadaptationlarge}
Edward~J. Hu, Yelong Shen, Phillip Wallis, Zeyuan Allen-Zhu, Yuanzhi Li, Shean
  Wang, Lu~Wang, and Weizhu Chen. 2021.
\newblock \href {https://arxiv.org/abs/2106.09685} {Lora: Low-rank adaptation
  of large language models}.
\newblock \emph{Preprint}, arXiv:2106.09685.

\bibitem[{Hu et~al.(2019)Hu, Chan, Liu, Zhao, Ma, and Yan}]{hu2019gsn}
Wenpeng Hu, Zhangming Chan, Bing Liu, Dongyan Zhao, Jinwen Ma, and Rui Yan.
  2019.
\newblock Gsn: A graph-structured network for multi-party dialogues.
\newblock \emph{arXiv preprint arXiv:1905.13637}.

\bibitem[{Lai et~al.(2024)Lai, Mesgar, and Fraser}]{lai2024llms}
Wen Lai, Mohsen Mesgar, and Alexander Fraser. 2024.
\newblock Llms beyond english: Scaling the multilingual capability of llms with
  cross-lingual feedback.
\newblock \emph{arXiv preprint arXiv:2406.01771}.

\bibitem[{Li et~al.(2024)Li, Huang, Ching, Dai, and
  Chen}]{li-etal-2024-prealign}
Jiahuan Li, Shujian Huang, Aarron Ching, Xinyu Dai, and Jiajun Chen. 2024.
\newblock \href {https://doi.org/10.18653/v1/2024.emnlp-main.572}
  {{P}re{A}lign: Boosting cross-lingual transfer by early establishment of
  multilingual alignment}.
\newblock In \emph{Proceedings of the 2024 Conference on Empirical Methods in
  Natural Language Processing}, pages 10246--10257, Miami, Florida, USA.
  Association for Computational Linguistics.

\bibitem[{Lin et~al.(2021)Lin, Liu, Winata, Cahyawijaya, Madotto, Bang, Ishii,
  and Fung}]{lin2021xpersona}
Zhaojiang Lin, Zihan Liu, Genta~Indra Winata, Samuel Cahyawijaya, Andrea
  Madotto, Yejin Bang, Etsuko Ishii, and Pascale Fung. 2021.
\newblock Xpersona: Evaluating multilingual personalized chatbot.
\newblock In \emph{Proceedings of the 3rd Workshop on Natural Language
  Processing for Conversational AI}, pages 102--112.

\bibitem[{Liu et~al.(2020)Liu, Gu, Goyal, Li, Edunov, Ghazvininejad, Lewis, and
  Zettlemoyer}]{liu-etal-2020-multilingual-denoising}
Yinhan Liu, Jiatao Gu, Naman Goyal, Xian Li, Sergey Edunov, Marjan
  Ghazvininejad, Mike Lewis, and Luke Zettlemoyer. 2020.
\newblock \href {https://doi.org/10.1162/tacl_a_00343} {Multilingual denoising
  pre-training for neural machine translation}.
\newblock \emph{Transactions of the Association for Computational Linguistics},
  8:726--742.

\bibitem[{Liu et~al.(2023)Liu, Nie, Cai, Wang, Niu, Zhang, Sachan, and
  Peng}]{liu2023xdailydialog}
Zeming Liu, Ping Nie, Jie Cai, Haifeng Wang, Zheng-Yu Niu, Peng Zhang, Mrinmaya
  Sachan, and Kaiping Peng. 2023.
\newblock Xdailydialog: A multilingual parallel dialogue corpus.
\newblock In \emph{Proceedings of the 61st Annual Meeting of the Association
  for Computational Linguistics (Volume 1: Long Papers)}, pages 12240--12253.

\bibitem[{Lowe et~al.(2015)Lowe, Pow, Serban, and Pineau}]{lowe2015ubuntu}
Ryan Lowe, Nissan Pow, Iulian Serban, and Joelle Pineau. 2015.
\newblock The ubuntu dialogue corpus: A large dataset for research in
  unstructured multi-turn dialogue systems.
\newblock \emph{arXiv preprint arXiv:1506.08909}.

\bibitem[{Muennighoff et~al.(2022)Muennighoff, Wang, Sutawika, Roberts,
  Biderman, Scao, Bari, Shen, Yong, Schoelkopf, Tang, Radev, Aji, Almubarak,
  Albanie, Alyafeai, Webson, Raff, and Raffel}]{muennighoff2022crosslingual}
Niklas Muennighoff, Thomas Wang, Lintang Sutawika, Adam Roberts, Stella
  Biderman, Teven~Le Scao, M~Saiful Bari, Sheng Shen, Zheng-Xin Yong, Hailey
  Schoelkopf, Xiangru Tang, Dragomir Radev, Alham~Fikri Aji, Khalid Almubarak,
  Samuel Albanie, Zaid Alyafeai, Albert Webson, Edward Raff, and Colin Raffel.
  2022.
\newblock \href {https://arxiv.org/abs/2211.01786} {Crosslingual generalization
  through multitask finetuning}.
\newblock \emph{Preprint}, arXiv:2211.01786.

\bibitem[{Mulcaire et~al.(2019)Mulcaire, Kasai, and
  Smith}]{mulcaire2019polyglot}
Phoebe Mulcaire, Jungo Kasai, and Noah~A Smith. 2019.
\newblock Polyglot contextual representations improve crosslingual transfer.
\newblock \emph{arXiv preprint arXiv:1902.09697}.

\bibitem[{Ponti et~al.(2020)Ponti, Glava{\v{s}}, Majewska, Liu, Vuli{\'c}, and
  Korhonen}]{ponti2020xcopa}
Edoardo~Maria Ponti, Goran Glava{\v{s}}, Olga Majewska, Qianchu Liu, Ivan
  Vuli{\'c}, and Anna Korhonen. 2020.
\newblock Xcopa: A multilingual dataset for causal commonsense reasoning.
\newblock In \emph{Proceedings of the 2020 Conference on Empirical Methods in
  Natural Language Processing (EMNLP)}, pages 2362--2376.

\bibitem[{Shmueli and Ku(2019)}]{shmueli2019socialnlp}
Boaz Shmueli and Lun-Wei Ku. 2019.
\newblock Socialnlp emotionx 2019 challenge overview: Predicting emotions in
  spoken dialogues and chats.
\newblock \emph{arXiv preprint arXiv:1909.07734}.

\bibitem[{Singh et~al.(2024)Singh, Patidar, and Vig}]{singh2024translating}
Pushpdeep Singh, Mayur Patidar, and Lovekesh Vig. 2024.
\newblock Translating across cultures: Llms for intralingual cultural
  adaptation.
\newblock In \emph{Proceedings of the 28th Conference on Computational Natural
  Language Learning}, pages 400--418.

\bibitem[{Touvron et~al.(2023)Touvron, Lavril, Izacard, Martinet, Lachaux,
  Lacroix, Rozi{\`e}re, Goyal, Hambro, Azhar et~al.}]{touvron2023llama}
Hugo Touvron, Thibaut Lavril, Gautier Izacard, Xavier Martinet, Marie-Anne
  Lachaux, Timoth{\'e}e Lacroix, Baptiste Rozi{\`e}re, Naman Goyal, Eric
  Hambro, Faisal Azhar, et~al. 2023.
\newblock Llama: Open and efficient foundation language models.
\newblock \emph{arXiv preprint arXiv:2302.13971}.

\bibitem[{Wang et~al.(2019)Wang, Mayhew, Roth et~al.}]{wang2019cross}
Zihan Wang, Stephen Mayhew, Dan Roth, et~al. 2019.
\newblock Cross-lingual ability of multilingual bert: An empirical study.
\newblock \emph{arXiv preprint arXiv:1912.07840}.

\bibitem[{Whitehouse et~al.(2023)Whitehouse, Choudhury, and
  Aji}]{whitehouse2023llm}
Chenxi Whitehouse, Monojit Choudhury, and Alham Aji. 2023.
\newblock Llm-powered data augmentation for enhanced cross-lingual performance.
\newblock In \emph{Proceedings of the 2023 Conference on Empirical Methods in
  Natural Language Processing}, pages 671--686.

\bibitem[{Xue et~al.(2021)Xue, Constant, Roberts, Kale, Al-Rfou, Siddhant,
  Barua, and Raffel}]{xue-etal-2021-mt5}
Linting Xue, Noah Constant, Adam Roberts, Mihir Kale, Rami Al-Rfou, Aditya
  Siddhant, Aditya Barua, and Colin Raffel. 2021.
\newblock \href {https://doi.org/10.18653/v1/2021.naacl-main.41} {m{T}5: A
  massively multilingual pre-trained text-to-text transformer}.
\newblock In \emph{Proceedings of the 2021 Conference of the North American
  Chapter of the Association for Computational Linguistics: Human Language
  Technologies}, pages 483--498, Online. Association for Computational
  Linguistics.

\bibitem[{Yang et~al.(2024)Yang, Yang, Hui, Zheng, Yu, Zhou, Li, Li, Liu,
  Huang, Dong, Wei, Lin, Tang, Wang, Yang, Tu, Zhang, Ma, Yang, Xu, Zhou, Bai,
  He, Lin, Dang, Lu, Chen, Yang, Li, Xue, Ni, Zhang, Wang, Peng, Men, Gao, Lin,
  Wang, Bai, Tan, Zhu, Li, Liu, Ge, Deng, Zhou, Ren, Zhang, Wei, Ren, Liu, Fan,
  Yao, Zhang, Wan, Chu, Liu, Cui, Zhang, Guo, and
  Fan}]{yang2024qwen2technicalreport}
An~Yang, Baosong Yang, Binyuan Hui, Bo~Zheng, Bowen Yu, Chang Zhou, Chengpeng
  Li, Chengyuan Li, Dayiheng Liu, Fei Huang, Guanting Dong, Haoran Wei, Huan
  Lin, Jialong Tang, Jialin Wang, Jian Yang, Jianhong Tu, Jianwei Zhang,
  Jianxin Ma, Jianxin Yang, Jin Xu, Jingren Zhou, Jinze Bai, Jinzheng He,
  Junyang Lin, Kai Dang, Keming Lu, Keqin Chen, Kexin Yang, Mei Li, Mingfeng
  Xue, Na~Ni, Pei Zhang, Peng Wang, Ru~Peng, Rui Men, Ruize Gao, Runji Lin,
  Shijie Wang, Shuai Bai, Sinan Tan, Tianhang Zhu, Tianhao Li, Tianyu Liu,
  Wenbin Ge, Xiaodong Deng, Xiaohuan Zhou, Xingzhang Ren, Xinyu Zhang, Xipin
  Wei, Xuancheng Ren, Xuejing Liu, Yang Fan, Yang Yao, Yichang Zhang, Yu~Wan,
  Yunfei Chu, Yuqiong Liu, Zeyu Cui, Zhenru Zhang, Zhifang Guo, and Zhihao Fan.
  2024.
\newblock \href {https://arxiv.org/abs/2407.10671} {Qwen2 technical report}.
\newblock \emph{Preprint}, arXiv:2407.10671.

\end{thebibliography}

\appendix

\section{Preliminary Experiments}\label{appendixa}
\begin{figure*}[!t]
	\centering
	\resizebox{\linewidth}{!}{ 
		\includegraphics[width=\textwidth]{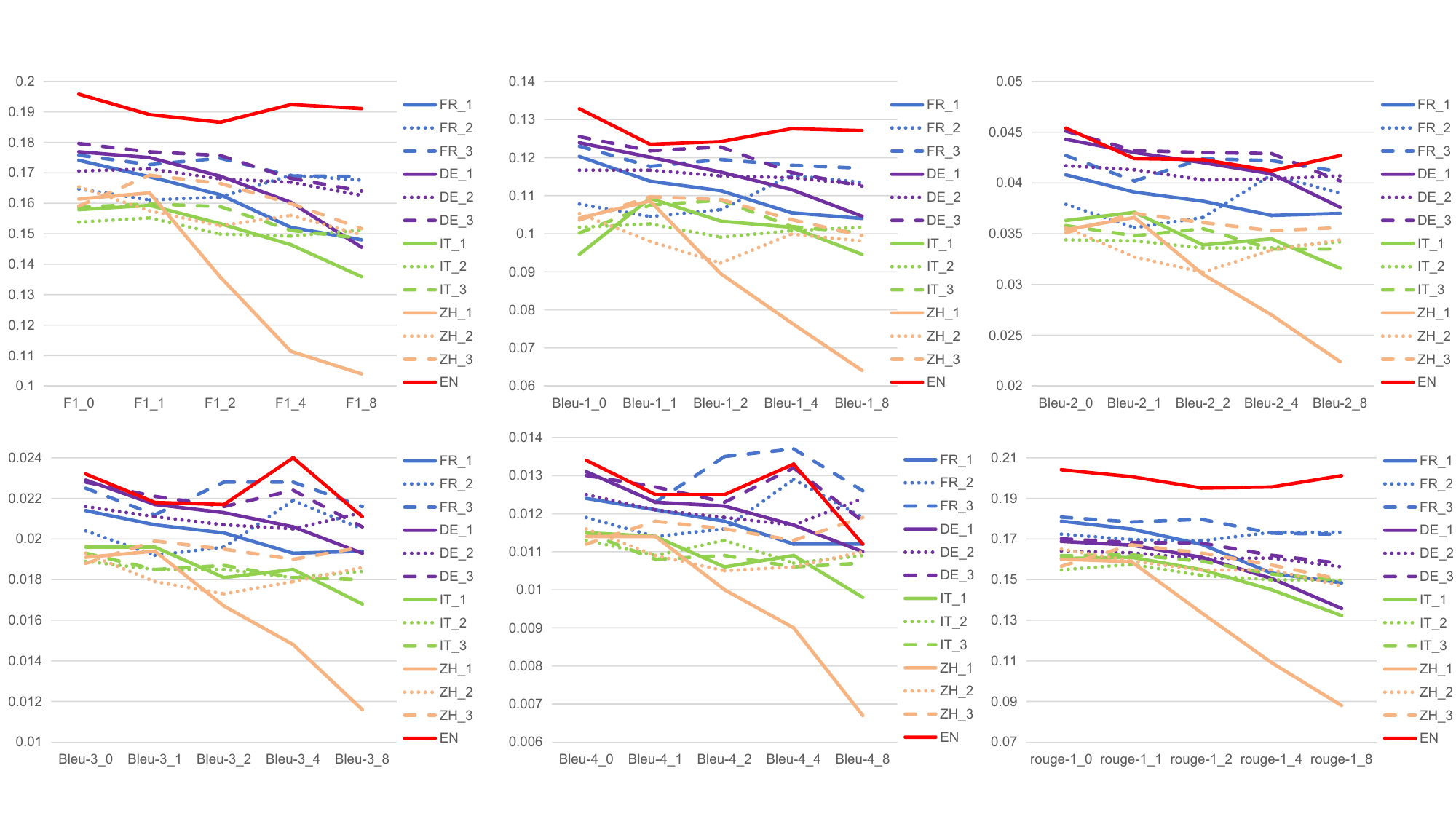}}
	\caption{The results of Preliminary Experiments. $\{metric\}\_{n}$ means n examples used in the ICL and $\{Language\}\_{x}$ means Type X template used.}
	\label{fig:pre_case}
\end{figure*}

Before constructing the full dataset, we first conducted a small-scale collection of approximately 1,000 English multi-party dialogue samples. Each sample contained at least three speakers and covered a variety of topics to ensure diversity and complexity in the dialogues. To evaluate the large model's multi-party dialogue response generation capability, we used In-Context Learning (ICL) testing. We found that as the number of examples in the ICL increased, there was virtually no improvement in the quality of the model's generated responses. This was particularly evident with the English data, which led us to question how the large model would perform on relatively sparse, under-pretrained corpora. To investigate this further, we expanded the 1,000 English samples to include French, German, Italian, and Chinese. The results are shown in Figure \ref{fig:pre_case}.
\begin{figure*}[!t]
	\centering
	\resizebox{\linewidth}{!}{ 
		\includegraphics[width=\textwidth]{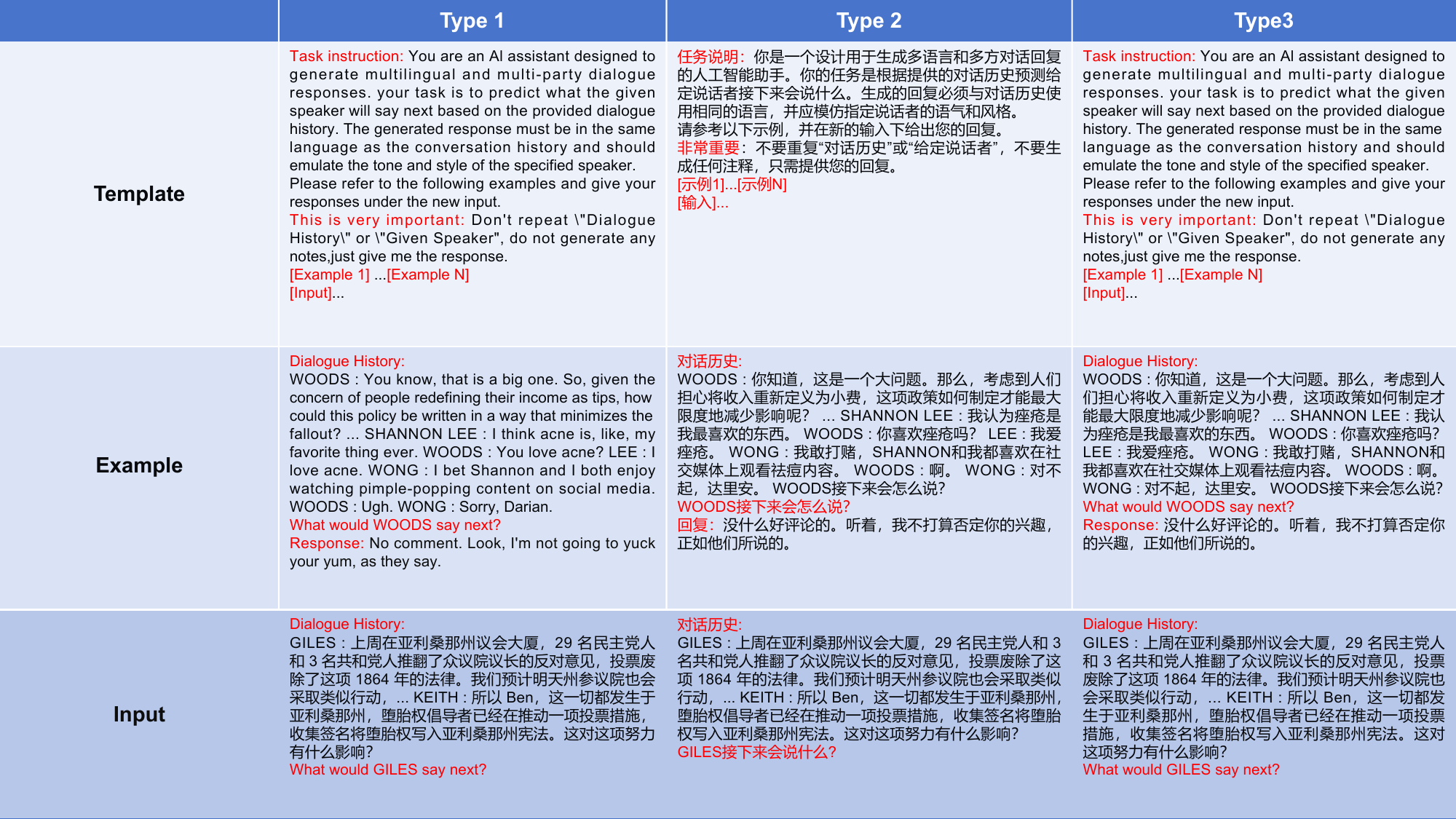}}
	\caption{This is an case with the target language in Chinese, showcasing three different template types (Type 1, Type 2, Type 3).}
	\label{fig:template}
\end{figure*}
Following the approach of \cite{choenni-etal-2023-languages} et al., we applied different templates combined with ICL for further validation. Specially, we designed three different experimental methods to validate the model's performance variations:
\begin{itemize}
	\item[1.] description Using English templates and English examples. In this setting, the model learns and infers based on English examples, effectively leveraging the provided data.
	
	\item[2.] Using target language templates and target language examples. For example, when testing with Chinese data, we employ Chinese templates and Chinese examples.
	
	\item[3.] Using English templates and target language examples.
\end{itemize}
We also experimented with various template formats, but the results were largely similar across different configurations. As a result, we selected one template format that was used in the final experiments, which is illustrated in the Figure \ref{fig:template}. Additionally, we tested the model's performance with different numbers of examples (0, 1, 2, 4, 8). 

The results are shown in Figure \ref{fig:pre_case}. Regardless of the template configurations used, we observed that as the number of examples in the ICL increased, the evaluation metrics generally decreased. These results raise important questions about the capabilities of LLMs and why ICL fails to lead to improvements in this context. LLMs are typically strong due to the vast amount of data they are trained on, which helps them learn conversational patterns and produce contextually appropriate responses. However, in the complex environment of multi-party dialogues, where several speakers interact in dynamic ways, the models' dependency on data-driven pattern matching falls short. They face challenges in effectively incorporating additional examples to enhance their understanding of the dialogue structure or adjust their reasoning to the specific context of the task.
\section{Data Collection and Preprocessing}\label{appendixb}
This chapter details the process of data collection, filtering, cleaning, and segmentation to construct our high-quality multilingual dataset.
\subsection{Data Collection}
We selected podcasts as our primary data source due to their ability to provide authentic conversational content. Most podcast episodes involve discussions among three or more participants, effectively simulating real-world multi-party interaction scenarios. Data was collected from publicly accessible podcast web: \href{www.npr.org}{www.npr.org}, including Life Kit, Pop Culture Happy Hour, Short Wave, Throughline, Consider This from NPR, It’s Been a Minute, The NPR Politics Podcast, The Lawfare Podcast, and Planet Money. All these programs are in the public domain, freely available for access and download.

The complete transcription content of the podcasts was fetched by constructing and sending HTTP requests to the relevant pages. HTML responses were parsed to extract transcription data. Specifically:
\begin{itemize}
	\item[1.] The HTML structure was parsed to extract the page titles.
	\item[2.] The \textbf{<div> }sections containing transcription text were identified and processed.
	\item[3.] Extraneous content, such as timestamps and bracketed text, was removed or converted into readable text.
	\item[4.] A regular expression \texttt{r'(\textbackslash w[\textbackslash w\textbackslash s,]*)\\:\textbackslash s*([\^{}:]*?)(?=\textbackslash w[\textbackslash w\textbackslash s,]*:|\$)'}was applied to separate "Speaker: Utterance" pairs, aligning speakers with their corresponding utterances.
\end{itemize}
\begin{table*}[!t]
	\centering
	\renewcommand\arraystretch{1.5}
	\resizebox{\linewidth}{!}{
		\begin{tabular}{|l|p{16cm}|}
			\hline
			\textbf{Type} & \textbf{Translated script} \\ \hline
			\textbf{Origin} & An untried producer and his cast of complete unknowns prepare to make television history while network suits breathe down their necks. And just about everything that can go wrong does, like, completely, totally. It's bad. I'm Aisha Harris. \\ \hline
			\textbf{Google} & \begin{CJK}{UTF8}{gbsn}一位未经验考验的制片人和他的一众无名小卒准备创造电视历史，而电视网络高层对他们紧追不舍。几乎所有可能出错的事情都完全出错了。这很糟糕。我是艾莎·哈里斯。\end{CJK} \\ \hline
			\textbf{Baidu} & \begin{CJK}{UTF8}{gbsn}一位未经验考验的制片人和他的一群完全未知的人准备创造电视历史，而网络服则压在他们的脖子上。几乎所有可能出错的事情会完全、彻底地发生。这很糟糕。我是艾莎·哈里斯。\end{CJK} \\ \hline
			\textbf{DeepL} & \begin{CJK}{UTF8}{gbsn}一位未经验考验的制片人和他的一帮完全不知名的演员准备创造电视历史，而电视台的西装革履让他们喘不过气来。一切可能出错的地方都出了错，就像，完全，彻底。这很糟糕。我是艾莎·哈里斯。\end{CJK} \\ \hline
			\textbf{iFlytek} & \begin{CJK}{UTF8}{gbsn}一位未经验考验的制片人和他的一帮完全不知名的演员准备创造电视历史，而电视台的西装束紧着他们。几乎所有可能出错的事情会发生，就像，完全，彻底。这很糟糕。我是艾莎·哈里斯。\end{CJK} \\ \hline
			\textbf{LLaMA3} & \begin{CJK}{UTF8}{gbsn}一位毫无经验的制片人和他全新演员阵容准备创造电视史，可电视台的高层却紧盯着他们。几乎所有可能出错的地方都会发生，真是糟透了。我是艾莎·哈里斯。\end{CJK} \\ \hline
			\textbf{Human} & \begin{CJK}{UTF8}{gbsn}一个毫无经验的制片人带领一群毫无名气的演员，准备改写电影史，可电视台高层却死死盯着他们的一举一动。结果几乎所有地方都出差错了，真是糟透了。我是艾莎·哈里斯。\end{CJK} \\ \hline
	\end{tabular}}
	\caption{Specific case comparisons of the translation tools.}
	\label{tab:translation_comparison}
\end{table*}
\subsection{Data Filtering}
Post extraction, a set of filtering rules was applied to eliminate noise from the data:
\begin{itemize}
	\item[1.] Platform Labels: Phrases such as ``Copyright,'' ``(Laughter),'' ``(SOUNDBITE OF TV SHOW, THE TONIGHT SHOW),'' and ``HOST, BYLINE'' were removed.
	\item[2.] URLs: Any URL strings within the text were removed.
\end{itemize}
\subsection{Data Segmentation}
Podcast episodes typically range from 80 to 100 utterances, making segmentation essential to generate manageable samples. The following rules were used:
\begin{itemize}
	\item[1.] Segment Length: Long conversations were split into shorter segments, ensuring each contained 7 to 15 utterances.
	\item[2.] Speaker Presence: The speaker generating the response must appear in the prior dialogue history to ensure coherence.
	\item[3.] Response Quality: Segments with overly short or long responses, or responses deemed meaningless (e.g., ``Yeah, wel'' or ``Alright''), were excluded.
\end{itemize}
\section{Comparison of Translate Method}\label{appendixc}
\subsection{Human Evaluation for Different Translation Methods}
\begin{table}[!t]
	\centering
	\renewcommand\arraystretch{1.2}
	\resizebox{\linewidth}{!}{ 
		\begin{tabular}{lccc}
			\hline
			\textbf{Translation Method} & \textbf{W-Level} & \textbf{U-Level} & \textbf{C-Level} \\ \hline
			Google Translate            & 92.6               & 86.9                     & 84.5                        \\
			Baidu Translate             & 88.2               & 73.1                     & 74.7                        \\
			DeepL Translate             & 88.7               & 70.8                     & 73.4                        \\
			iFLYTEK Translate           & 86.2               & 70.2                     & 72.3                        \\
			LLaMA 3.1                     & 90.4               & 80.3                     & 82.8                        \\
			ChatGPT-4o                     & 94.5               & 85.9                     & 87.8                        \\ \midrule
			Human Translation           & \textbf{99.2}               & \textbf{98.5}                     & \textbf{95.4}                        \\ \hline
	\end{tabular}}
	\caption{Evaluation Results of Translation Methods. W-Level, U-Level and C-Level refer to Word-Level, Utterance-Level and Conversation-Level.}
	\label{table:translation_evaluation}
\end{table}
\begin{table}[ht]
	\centering
		\resizebox{\linewidth}{!}{ 
	\begin{tabular}{llcccc}
		\toprule
		\textbf{Language} & \textbf{Method} & \textbf{F1} & \textbf{B-1} & \textbf{B-2} & \textbf{R-L} \\
		\midrule
		Chinese & GPT-4o           & 17.45 & 10.93 & 4.59 & 15.01 \\
		Chinese & Google & 17.61 & 11.01 & 4.55 & 14.97 \\
		German  & GPT-4o           & 16.97 & 11.60 & 4.91 & 13.88 \\
		German  & Google & 17.12 & 11.51 & 4.95 & 13.84 \\
		\bottomrule
	\end{tabular}}
	\caption{Performance comparison of LLaMA 3.1 trained on translations generated by GPT-4o vs. Google Translate (5k samples each).}
	\label{tab:gpt4o_vs_googletranslate}
\end{table}
For human evaluation, we began by selecting 50 representative samples that covered a diverse range of topics (e.g., law, politics, daily life), dialogue turns (ranging from 7 to 15 turns), and multi-party podcast segments. Each method translated the English transcripts into five other languages (Chinese, Japanese, French, German, and Italian), resulting in 300 translations. To ensure objectivity and reliability, three linguistics experts reviewed each translation in multiple rounds. The review process identified issues such as lexical misusage, semantic distortion, or contextual inconsistency in the translated outputs. Ultimately, the evaluation focused on three dimensions:
\begin{itemize}
	\item \textbf{Word-level} evaluation assessed the accuracy of individual word translations and checked for spelling errors.
	\item \textbf{Utterance-level} evaluation examined sentence fluency, semantic accuracy, and appropriateness in context.
	\item \textbf{Conversation-level} evaluation ensured contextual consistency and alignment with the original dataset.
\end{itemize}
As shown in Table \ref{table:translation_evaluation}, among Google Translate, DeepL, Baidu, iFLYTEK, LLaMA 3, ChatGPt-4o, both ChatGPT-4 and Google Translate have shown impressive results, with ChatGPT-4o slightly outperforming Google Translate. However, ChatGPT-4o requires careful handling of templates, as it may provide natural translations for some examples but sometimes offers literal translations that overlook specific contexts. \textbf{Considering translation speed and cost-effectiveness}, we opted for Google Translate as the more practical choice for large-scale multilingual tasks, just like prior works \cite{lai2024llms,liu2023xdailydialog}.
\subsection{Automatic Evaluation for Different Translation Methods}
We further validated translation quality through a small-scale generation experiment, as shown in Table \ref{tab:gpt4o_vs_googletranslate}. A subset of approximately 5,000 English samples was translated into Chinese and German using both GPT-4o and Google Translate. We then fine-tuned LLaMA 3.1 using each translated dataset and compared dialogue generation performance. Results showed minimal performance differences between the two translation sources. These findings support the use of Google Translate as a practical and reliable solution for large-scale multilingual dialogue construction.
\subsection{Case comparisons}
Specific case comparisons of the translation tools are detailed below.
\begin{table*}[!t]
	\centering
	\renewcommand\arraystretch{1.5}
	\resizebox{\linewidth}{!}{
		\begin{tabular}{|p{6cm}|p{6cm}|p{6cm}|}
			\hline
			\textbf{Origin}& \textbf{Prompt 1:}\begin{CJK}{UTF8}{gbsn}翻译成中文\end{CJK} & \textbf{Prompt 2:}\begin{CJK}{UTF8}{gbsn}翻译成中文，请不要直译，在这句话中有很多需要理解并意译的地方，请注意\end{CJK} \\
			\hline
			An untried producer and his cast of complete unknowns prepare to make television history while network suits breathe down their necks. And just about everything that can go wrong does, like, completely, totally. It's bad. I'm Aisha Harris. & \begin{CJK}{UTF8}{gbsn}一位未经试验的制片人和他的完全不为人知的演员阵容准备创造电视历史，而电视网络的高层正对他们施加巨大压力。几乎所有可能出错的事情都出了错，完全、彻底地错了。情况非常糟糕。我是Aisha Harris。 \end{CJK}
			& \begin{CJK}{UTF8}{gbsn}一位经验不足的制片人和一群完全没有名气的演员，正准备创造电视历史，而电视台的高层则在背后施压。几乎所有可能出错的事都发生了，而且是彻底的、完全的错误，局面一团糟。我是Aisha Harris。\end{CJK}\\
			\hline
	\end{tabular}}	

	\caption{This is the translation result by ChatGPT-4o. As you can see, using different prompts leads to completely different translations. The result with prompt 2 is much better than result with prompt 1.}
		\label{tab:chatgpt}
\end{table*}
To further illustrate our reasoning for choosing the Google Translate API, the Table \ref{tab:translation_comparison} and Table \ref{tab:chatgpt} present an actual translation snippet of a dialogue. In terms of word-level evaluation, all the translation APIs demonstrated accurate spelling without errors. Google's translation of "cast of complete unknowns" as "\begin{CJK}{UTF8}{gbsn}
一众无名小卒\end{CJK}" is both vivid and precise. In contrast, Baidu's "\begin{CJK}{UTF8}{gbsn}网络服装\end{CJK}" is a literal translation of "network suits," failing to capture the intended meaning. Similarly, DeepL's "\begin{CJK}{UTF8}{gbsn}西装革履\end{CJK}" and iFlytek's "\begin{CJK}{UTF8}{gbsn}西装\end{CJK}" translations of "network suits" are overly literal, lacking the semantic nuance of referring to "network executives."

For utterance-level evaluation, Google's rendering of "\begin{CJK}{UTF8}{gbsn}电视网络高层对他们紧追不舍\end{CJK}" effectively conveys the urgency in "network suits breathe down their necks," while maintaining fluency and accuracy in alignment with the original context. Baidu's translation of "\begin{CJK}{UTF8}{gbsn}一位未经尝试的制片人\end{CJK}" is less precise and deviates from the original meaning. The phrase "\begin{CJK}{UTF8}{gbsn}完全、彻底地发生\end{CJK}" is grammatically correct but verbose and unnatural. Overall, Baidu's translation style tends to be rigid, lacking smooth and natural expression. DeepL and iFlytek's translations, such as "\begin{CJK}{UTF8}{gbsn}就像，完全，彻底\end{CJK}," fail to accurately reflect the overall semantics and lack fluency. Meanwhile, LLaMA3's translation of "cast of complete unknowns" as "\begin{CJK}{UTF8}{gbsn}全新演员阵容\end{CJK}" shifts the original "unknown" connotation to a more positive tone, which is inconsistent with the original context.

For conversation-level evaluation, Google Translate demonstrates coherence and effectively conveys the original meaning with high parallelism, maintaining a natural conversational tone. While LLaMA3 generally performs well in contextual consistency, certain examples, such as this one, show occasional mismatches in contextual understanding.

Comprehensive Evaluation: The Google Translate API outperforms other tools and models by better capturing the urgency and fluency of the original text. Although there are still slight deficiencies compared to human translation, it is sufficiently effective for large-scale multilingual parallel data construction. While LLaMA3 translations also exhibit reasonable quality, they sometimes produce contextually irrelevant outputs or fail to handle languages with scarce training data effectively, generating extraneous or unfaithful translations.
\section{Dataset Quality: Detailed evaluation criteria and the scoring results}\label{appendixd}
To assess the quality of the dataset, we employed expert evaluation. Specifically, we randomly selected 1000 conversations (200 samples for each language), covering various topics. The review was conducted based on the following aspects, with scores ranging from 1 to 5 for each metric. 
\paragraph{Cross-lingual Consistency Analysis}
	use the Scoring Criteria below to assess how well the target-language text preserves the semantic meaning and conceptual structure of the English source text.
	\begin{itemize}
		\item Semantic Equivalence: Accuracy in capturing key information, terminology, and meaning from the source.
		
		\item Conceptual Alignment: Faithfulness to the overall theme, reasoning chain, and logical flow of the source text.
	\end{itemize}
The Scoring Criteria and results are shown in Table \ref{tab:translation_evaluation} and Table \ref{tab:translation_quality}.
\paragraph{Linguistic and Cultural Consistency}
	focuses on the quality of the target-language text in its own linguistic and cultural context.
	\begin{itemize}
		 \item Naturalness: Whether sentences are fluent, grammatically sound, and idiomatic.

		\item Cultural Appropriateness: Alignment with cultural norms, conventions, and stylistic preferences.
	\end{itemize}
The Scoring Criteria and  result are shown in Table \ref{tab:naturalness_culture} and Table \ref{tab:naturalness_scores}.
\begin{table*}[ht]
	\centering
	\begin{tabular}{|c|p{6cm}|p{6.5cm}|}
		\hline
		\textbf{Score} & \textbf{Semantic Equivalence} & \textbf{Conceptual Alignment} \\
		\hline
		1 & Many key details lost or mistranslated; Significant deviation from source meaning & Main topic is unclear or completely misaligned; Severe logical gaps or missing core concepts \\
		\hline
		2 & Substantial inaccuracies in important information; Inadequate handling of key terms & Overall theme is partially discernible, but reasoning steps are shaky; Noticeable gaps in coherence \\
		\hline
		3 & Primary content is mostly correct; Some inaccuracies in less crucial details & Main idea is largely maintained; Minor logical jumps but generally comprehensible \\
		\hline
		4 & Nearly all key information is accurately preserved; Only minor misinterpretations & The text closely follows the source theme; Reasoning and structure are coherent, with few gaps \\
		\hline
		5 & All core information perfectly conveyed; No noticeable semantic errors & Fully aligned with source logic; Clear and consistent reasoning; Nearly flawless \\
		\hline
	\end{tabular}
	\caption{Semantic Equivalence and Conceptual Alignment Scoring Guidelines}
	\label{tab:translation_evaluation}
\end{table*}
\begin{table*}[ht]
	\centering
	\begin{tabular}{|l|c|c|c|}
		\hline
		\textbf{Target Language} & \textbf{Samples} & \textbf{Semantic Equivalence} & \textbf{Conceptual Alignment} \\
		\hline
		ZH (Chinese) & 200 & 4.13 & 4.02 \\
		JA (Japanese) & 200 & 4.17 & 4.11 \\
		FR (French) & 200 & 4.32 & 4.25 \\
		DE (German) & 200 & 4.05 & 3.92 \\
		IT (Italian) & 200 & 3.93 & 3.88 \\
		\hline
	\end{tabular}
	\caption{Quality Scores by Language}
	\label{tab:translation_quality}
\end{table*}
\begin{table*}[ht]
	\centering
	\begin{tabular}{|c|p{6.5cm}|p{6.5cm}|}
		\hline
		\textbf{Score} & \textbf{Naturalness} & \textbf{Cultural Appropriateness} \\
		\hline
		1 & Numerous grammatical or lexical errors. Difficult to read or parse & Severely mismatched with target-language cultural norms. Potentially offensive or highly unnatural expressions \\
		\hline
		2 & Limited fluency; awkward or clumsy phrasing; Noticeable "translationese" & Certain mismatches in cultural context; Expressions that appear slightly off or inauthentic \\
		\hline
		3 & Mostly coherent sentences; Minor issues with word choice or syntax & Understandable but not fully adapted to local norms; Some cultural references may be vague or misplaced \\
		\hline
		4 & Generally fluent and easy to follow; Few or negligible foreign constructs & Largely compatible with the target culture; Stylistically appropriate in most contexts \\
		\hline
		5 & Exceptionally natural, no apparent translation artifacts; Flawless grammar & Perfect cultural fit; expressions and style feel native; No inaccuracies or incongruities in cultural references \\
		\hline
	\end{tabular}
	\caption{Scoring Criteria for Naturalness and Cultural Appropriateness}
	\label{tab:naturalness_culture}
\end{table*}
\begin{table*}[ht]
	\centering
	\begin{tabular}{|l|c|c|c|}
		\hline
		\textbf{Target Language} & \textbf{\#Samples} & \textbf{Naturalness} & \textbf{Cultural Appropriateness} \\
		\hline
		ZH (Chinese) & 200 & 3.94 & 4.10 \\
		JA (Japanese) & 200 & 4.03 & 4.12 \\
		FR (French)   & 200 & 4.25 & 4.37 \\
		DE (German)   & 200 & 3.88 & 3.96 \\
		IT (Italian)  & 200 & 3.72 & 3.85 \\
		\hline
	\end{tabular}
	\caption{Naturalness and Cultural Appropriateness Scores by Language}
	\label{tab:naturalness_scores}
\end{table*}

\section{Detailed training and inference command}\label{appendixe}
    \begin{itemize}
    	\item     Training qwen-2.5 7b: \begin{lstlisting}
llamafactory-cli train --stage sft --do_train --model_name_or_path path_to/Qwen2.5-7B-Instruct --dataset train_en --dataset_dir path_to_your_dataset --template qwen --finetuning_type lora --lora_target q_proj,v_proj --output_dir path_to_save_dir --overwrite_cache --overwrite_output_dir --cutoff_len 2048 --preprocessing_num_workers 16 --per_device_train_batch_size 2 --per_device_eval_batch_size 1 --gradient_accumulation_steps 4 --lr_scheduler_type cosine --logging_steps 50 --warmup_steps 200 --save_steps 1000 --eval_steps 1000 --evaluation_strategy steps --learning_rate 5e-5 --num_train_epochs 2.0 --val_size 0.1 --plot_loss --fp16\end{lstlisting}
		\item     Inferring qwen-2.5 7b: \begin{lstlisting}
llamafactory-cli train --stage sft --do_predict --model_name_or_path path_to/Qwen2.5-7B-Instruct  --adapter_name_or_path path_to_save_dir --finetuning_type lora --dataset_dir path_to_your_dataset --template qwen --finetuning_type lora --lora_target q_proj,v_proj --output_dir output_dir --overwrite_cache --overwrite_output_dir --cutoff_len 2048 --preprocessing_num_workers 16 --per_device_eval_batch_size 4 --predict_with_generate --eval_dataset test_en\end{lstlisting}
		\item     Training llama-3.1 8b: \begin{lstlisting}
llamafactory-cli train --stage sft --do_train --model_name_or_path path_to/Meta-Llama-3.1-8B-Instruct --dataset train_en --dataset_dir path_to_your_dataset --template llama3 --finetuning_type lora --lora_target q_proj,v_proj --output_dir path_to_save_dir --overwrite_cache --overwrite_output_dir --cutoff_len 2048 --preprocessing_num_workers 16 --per_device_train_batch_size 2 --per_device_eval_batch_size 1 --gradient_accumulation_steps 4 --lr_scheduler_type cosine --logging_steps 50 --warmup_steps 200 --save_steps 1000 --eval_steps 1000 --evaluation_strategy steps --learning_rate 5e-5 --num_train_epochs 2.0 --val_size 0.1 --plot_loss --fp16\end{lstlisting}
		\item     Inferring llama-3.1 8b: \begin{lstlisting}
llamafactory-cli train --stage sft --do_predict --model_name_or_path path_to/Meta-Llama-3.1-8B-Instruct  --adapter_name_or_path path_to_save_dir --finetuning_type lora --dataset_dir path_to_your_dataset --template llama3 --finetuning_type lora --lora_target q_proj,v_proj --output_dir output_dir --overwrite_cache --overwrite_output_dir --cutoff_len 2048 --preprocessing_num_workers 16 --per_device_eval_batch_size 4 --predict_with_generate --eval_dataset test_en\end{lstlisting}
		\item     Training llama-3.1 70b:\begin{lstlisting}
llamafactory-cli train --stage sft --do_train --model_name_or_path path_to/Meta-Llama-3.1-70B-Instruct --dataset train_en --dataset_dir path_to_your_dataset --template llama3 --finetuning_type lora --lora_target q_proj,v_proj --output_dir path_to_save_dir --overwrite_cache --overwrite_output_dir --cutoff_len 2048 --preprocessing_num_workers 16 --per_device_train_batch_size 2 --per_device_eval_batch_size 1 --gradient_accumulation_steps 4 --lr_scheduler_type cosine --logging_steps 50 --warmup_steps 200 --save_steps 1000 --eval_steps 1000 --evaluation_strategy steps --learning_rate 5e-5 --num_train_epochs 2.0 --val_size 0.1 --plot_loss --fp16\end{lstlisting}
		\item     Inferring llama-3.1 70b: \begin{lstlisting}
python vllm_infer.py --model_name_or_path path_to/Meta-Llama-3.1-70B-Instruct --dataset test_en --dataset_dir path_to_your_dataset --template llama3 --adapter_name_or_path path_to_save_dir\end{lstlisting}
    \end{itemize}

\end{document}